\documentclass[lettersize,journal]{IEEEtran}
\usepackage{amsmath,amsfonts}
\usepackage{array}
\usepackage[caption=false,font=normalsize,labelfont=sf,textfont=sf]{subfig}
\usepackage{textcomp}
\usepackage{stfloats}
\usepackage{url}
\usepackage{verbatim}
\usepackage{graphicx}
\usepackage{cite}
\graphicspath{ {Images/} } 

\newcommand{\real}{\mbox{\rm I$\!$R}}

\newcommand{\bld}[1]{\mbox{\boldmath $#1$}} 

\newcommand{\F}{\bld{F}}

\newcommand{\M}{\bld{M}}

\usepackage[ruled,vlined,lined,linesnumbered,english]{algorithm2e}

\SetCommentSty{mycommfont}
\SetKwProg{Fn}{Function}{}{}

\begin{document}

\title{Mechanic Modeling and Nonlinear Optimal Control of Actively Articulated Suspension of Mobile Heavy-Duty Manipulators}

\author{Alvaro Paz, Jouni Mattila
\thanks{This work was supported by the Business Finland partnership project “Future all-electric rough terrain autonomous mobile manipulators” (Grant \#2334/31/2022). Corresponding author: Alvaro Paz alvaro.pazanaya@tuni.fi}
\thanks{All authors are with the Faculty of Engineering and Natural Sciences, Tampere University, Tampere, Finland}}



\maketitle

\begin{abstract}
This paper presents the analytic modeling of mobile heavy-duty manipulators with actively articulated suspension and its optimal control to maximize its static and dynamic stabilization. By adopting the screw theory formalism, we consider the suspension mechanism as a rigid multibody composed of two closed kinematic chains. This mechanical modeling allows us to compute the spatial inertial parameters of the whole platform as a function of the suspension's linear actuators through the articulated-body inertia method. Our solution enhances the computation accuracy of the wheels' reaction normal forces by providing an exact solution for the center of mass and inertia tensor of the mobile manipulator. Moreover, these inertial parameters and the normal forces are used to define metrics of both static and dynamic stability of the mobile manipulator and formulate a nonlinear programming problem that optimizes such metrics to generate an optimal stability motion that prevents the platform's overturning, such optimal position of the actuator is tracked with a state-feedback hydraulic valve control. We demonstrate our method's efficiency in terms of C++ computational speed, accuracy and performance improvement by simulating a 7 degrees-of-freedom heavy-duty parallel-serial mobile manipulator with four wheels and actively articulated suspension.
\end{abstract}


\section{Introduction}
\begin{figure}[t!] 
	\centering
	\includegraphics[trim={0.0cm 0.0cm 0.0cm 0.0cm},clip,width=6cm]{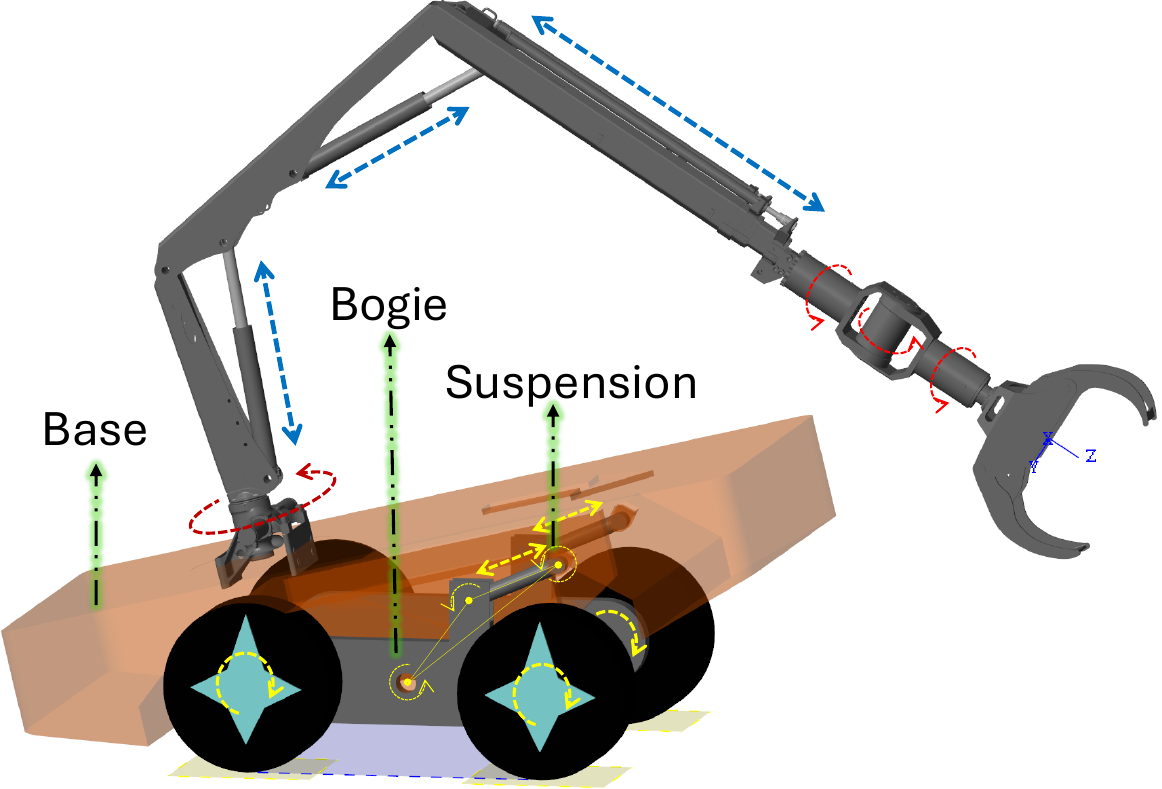}
	\vspace*{0cm}
	\caption{{\footnotesize {\bf Heavy-duty mobile manipulator.} This consists of a heavy-duty parallel-serial manipulator mounted on top of a wheeled mobile platform with actively articulated suspension powered by linear actuators.}}
	\label{fig:platform}
\end{figure}

\IEEEPARstart{M}{obile} heavy-duty machinery is widely applied and plays a critical role in industrial fields, such as construction, exploration, foresting, and mining. However the modeling and control of such bulky mechanisms \cite{RAMLI20171,mattila2017survey,abdel2003dynamics,johns2023framework} remain under-explored due to the new innovative features that can be endowed to them. Examples include the new generation of actuators \cite{bahari2024system} and sensors and new topological kinematic structures that need to be adequately addressed. Mobile manipulators (Fig. \ref{fig:platform}) are one of the most popular heavy-duty machinery type. A mobile manipulator is composed of a floating-base mobile platform commonly supported on wheels and equipped with an industrial manipulator on top, which has the ability to cope with high-payload tasks. Mobile robots need to maintain a mechanical static-and-dynamic equilibrium or stabilization that prevents them from falling down or overturning, particularly when traversing rough terrains, where ground inclination plays an important role, or in the case of heavy-duty machinery where an overturning event can result in serious injury. Multiple approaches have been investigated to generate feasible criteria that warranty an overtuning-free traversing. In addition, the mobile platform can be equipped with actuated-articulated suspensions, where its appropriate motion control can help mitigate the chances of overturning.
\subsection{Background and Context}
The dynamics of robot manipulators have been extensively studied through geometric \cite{lynch2017modern} and algebraic approaches \cite{Bib:Featherstone} by exploiting the inherent recursivity of rigid-body computations. The resulting recursive algorithms have straightforward implementation when considering tree-topology kinematic chains. However, a heavy-duty parallel-serial manipulator can contain not only serial kinematic chains in its topology but also closed kinematic chains \cite{zhu2010virtual}. Thus, the forward and inverse dynamics of such structures require tailored algorithms to retrieve analytical solutions of the closed-loops dynamics \cite{petrovic2022mathematical}\cite{alvaro2024analytical}.

The generation of dynamically feasible movements for the trajectory optimization of mobile robots remains challenging \cite{Bib:Bryson,posa2014direct,Bib:Betts}. Multiple approaches have been proposed to explore the static and dynamic metrics to monitor the mechanical stabilization during motion to prevent fall down or overturn. One of these approaches is center-of-mass formulations \cite{kajita2008legged,paz2024humanoid} for legged robots, where the center-of-mass orthogonal projection onto the terrain is forced to be inside the polygon of support. Other dynamical criteria are based on regularizing the zero moment point \cite{vukobratovic2004zero} and the robot's spatial momentum and centroidal momentum \cite{Bib:orin2013centroidal}. Additionally, an analytic solution for the normal reaction forces in wheeled robots has been reported \cite{petrovic2022analytic} as a good metric for overturning prevention.

The modeling and control of actively articulated suspensions of the mobile platform have been reported for center-of-mass-based stability criteria \cite{iagnemma2003control} and for normal-forces-based criteria \cite{hutter2015towards}. Both approaches successfully separated the static and dynamic stabilization by leveraging highly kinematically redundant wheeled suspensions.
\subsection{Paper Contributions}
The normal ground reaction forces for heavy-duty manipulators were analytically solved by \cite{petrovic2022analytic}, where the solution was bound by eight assumptions and many of the kinematic and inertial terms remained constant since the mobile platform was assumed to be a single rigid body. In this study we take into account the actuated-articulated suspension that generates a rigid multibody system with closed kinematic chains in its topology. Because of this consideration, many terms in the computation of normal forces become functions of the suspension actuators (see Section \ref{sec:forces} for a list of these varying terms). Considering this multibody suspension, the accuracy of the normal forces computation is significantly enhanced, as shown in Figs. \ref{fig:forces_k} and \ref{fig:forces_error}.

The control of actively articulated suspensions for highly articulated redundant floating bases was studied in \cite{iagnemma2003control}, \cite{hutter2015towards}. In particular, the quasi-static approach reported by \cite{iagnemma2003control} was reasonably justified by the very slow velocity mechanics of planetary exploration wheeled robots, while \cite{hutter2015towards} reported a more dynamic approach by considering the normal forces. Unlike the redundant mechanisms in \cite{iagnemma2003control}, \cite{hutter2015towards}, our work addresses the critical challenge of a nonredundant suspension mechanism where the lack of degrees of freedom (DOFs) represents a problem to be overcome when stabilizing the mobile robot. For instance, we combine in a single multiobjective nonlinear optimal control static and dynamic criteria same as those adopted in \cite{iagnemma2003control} and \cite{hutter2015towards} and successfully generate optimal suspension actuator commands that stabilize the mobile platform. The significant performance of our optimal control is depicted in Figs. \ref{fig:forces_control} and \ref{fig:optimal}. Moreover, a dynamic optimal normal forces distribution metric is shown in Fig. \ref{fig:diff_forces} and that of the quasi-static metric is shown in Fig. \ref{fig:com}, where both criteria are optimized into the same multiobjective cost function. In addition, unlike \cite{iagnemma2003control}, we consider optimal velocity and acceleration consistency for the suspension actuators and include smooth joint limits to avoid the piston effect of mechanical shock. We also show how these optimal actuator position can be used as a reference trajectory into a state-feedback hydraulic valve control which provides a better insight for future online experiments for an actively suspension hydraulically powered.

Using Lie groups and their algebras, we explicitly provide analytic solutions to the closed kinematic chains in the suspension. We demonstrate that this class of closed loops possesses the same kinematic topology as the one-DOF four-joint mechanism analyzed in \cite{zhu2010virtual}, \cite{petrovic2022mathematical}, \cite{alvaro2024analytical}. This analytic solution is then combined with the articulated-body inertia method \cite{Bib:Featherstone} for the assembling spatial inertia matrices of the whole mechanism and providing analytical derivations of its inertial parameters, which enhance the accuracy of normal force computation of the wheels. Moreover, we report through TABLE \ref{tab:table1} the computational times of our algorithms in a C++ implementation. These competitive computational times make our analytic solutions suitable for future real-time online implementations of nonlinear model predictive controls for overturning-free trajectory generation.

The rest of this paper is organized as follows. Section \ref{sec:mech} presents the topological structure of the entire platform, the heavy-duty manipulator's dynamics, and the second-order kinematics of the suspension's closed kinematic chains. The results are combined to assemble spatial inertias to retrieve analytic solutions for the inertial parameters that feed the wheels' normal force calculation. In Section \ref{sec:opt}, the spatial inertias and normal forces are used to generate metrics to define criteria for the multiobjective constrained optimal control. Section \ref{sec:results} presents the simulation results for the accuracy of normal force computation and performance of the optimal control in terms of center-of-mass regulation and normal force distribution. Finally, the conclusions and some considerations are presented.

\section{Heavy-Duty Mobile Manipulator Mechanics}
\label{sec:mech}
\begin{figure}[h!] 
	\centering
	\includegraphics[trim={0cm 0cm 0cm 0.0cm},clip,width=6cm]{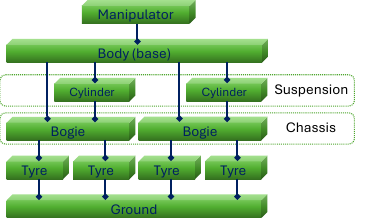}
	\vspace*{0cm}
	\caption{{\footnotesize Graphical representation of the topological structure.}}
	\label{fig:graph}
\end{figure}

The mobile manipulator with articulated suspension, depicted in Fig. \ref{fig:platform}, is a rigid multibody mechanism. Its kinematic topology is shown in Fig. \ref{fig:graph}, where a heavy-duty parallel-serial manipulator is mounted on top of the base rigid body. This base is parallelly connected to the chassis through two closed kinematic chains as suspension. Each closed chain contains a cylindrical linear actuator that is connected to a bogie of the chassis. Then, each bogie is connected to two wheels.

\begin{figure*}[h!]
	\centering
	\begin{minipage}[b]{0.32\textwidth}
		\includegraphics[trim={0cm 0cm 0cm 0cm},clip,width=\textwidth]{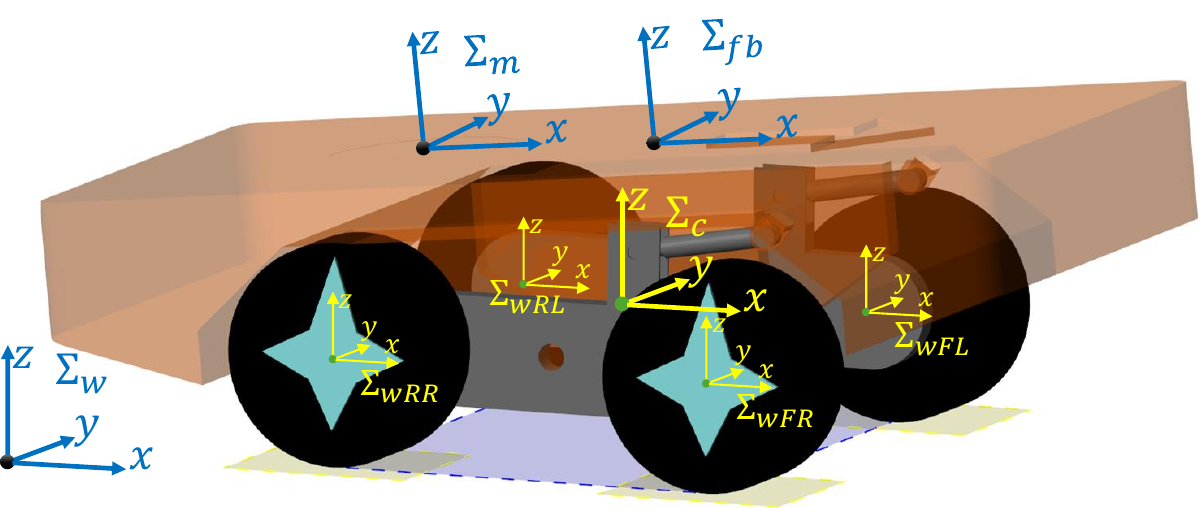}
		\label{fig:mani1}
	\end{minipage}
	\begin{minipage}[b]{0.32\textwidth}
		\includegraphics[trim={0cm 0cm 0cm 0cm},clip,width=\textwidth]{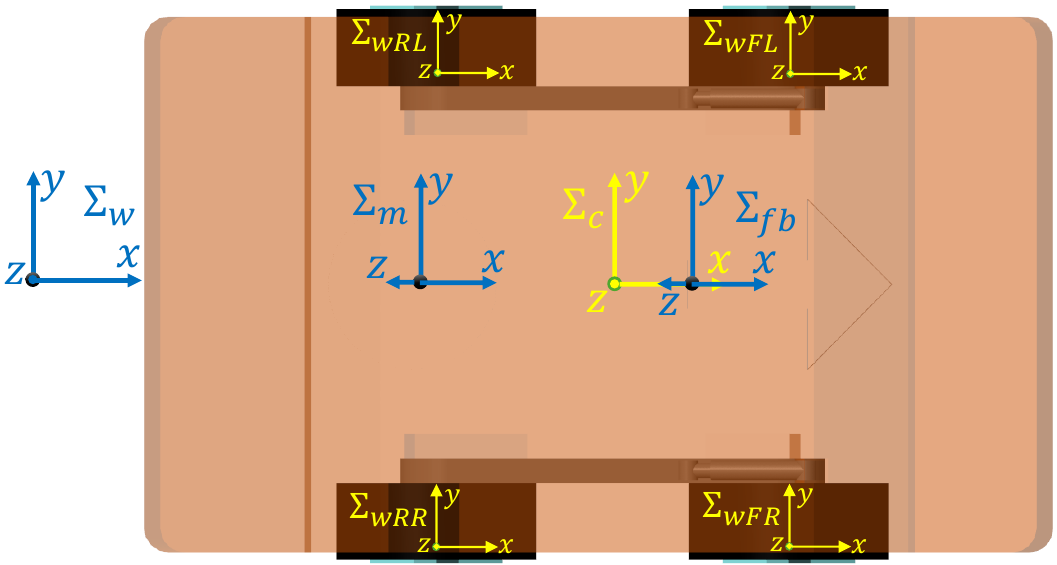}
		\label{fig:mani2}
	\end{minipage}
	\begin{minipage}[b]{0.32\textwidth}
		\includegraphics[trim={0cm 0cm 0cm 0cm},clip,width=\textwidth]{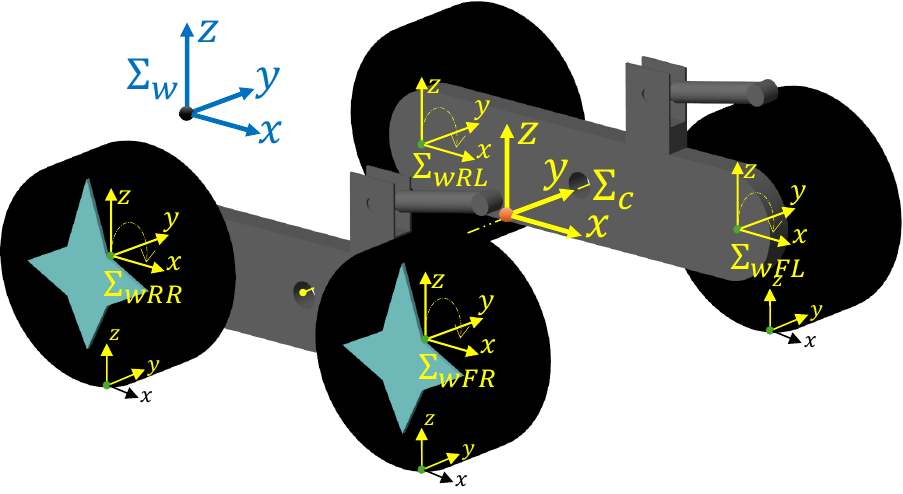}
		\label{fig:mani3}
	\end{minipage}
	\vspace*{-0.4cm}
	\caption{{\footnotesize {\bf Snapshots of the mobile platform with articulated suspension.} From left to right, the side view, top view, and chassis with wheels. The inertial reference frame of the world is represented by $\Sigma_{w}$ and the chassis reference frame by $\Sigma_{c}$. The manipulator reference frame $\Sigma_{m}$ and the base frame $\Sigma_{fb}$ are fixed to the base rigid body in red. The wheels' reference frames are depicted by $\Sigma_{wFR}$, $\Sigma_{wFL}$, $\Sigma_{wRR}$, and $\Sigma_{wRL}$.}}
	\label{fig:snaps}
\end{figure*}

The most important reference frames of the wheeled platform are shown in Fig. \ref{fig:snaps}, where $\Sigma_{w}$ is the world's inertial reference frame. The kinematic relation between $\Sigma_{w}$ and $\Sigma_{fb}$ is modeled as a 6-DoF floating base with configurational vector position $\bld{q}_{\!fb}\in\real^{6}$. By adopting the screw theory formalism \cite{lynch2017modern}, the forward kinematic of $\Sigma_{fb}$ with respect to $\Sigma_{w}$ is given by the transformation matrix $\bld{G}_{\!w}^{fb}(\bld{q}_{\!fb})$, which is a representation of an element of the Lie group $SE(3)$ and can be recursively computed using the product of exponentials introduced in \cite{brockett1984} and the exponential map \cite{Bib:selig}:
\begin{equation}
	\bld{G} \ = \ e^{[\bld{s}]q}
	\label{eq:op00}
\end{equation}
where $\bld{G}$ is a local transformation matrix, $q\in\real$ is the joint position, $\bld{s}$ is a screw vector of the Lie algebra $se(3)$, and $[\bld{s}]$ is its matrix form. The transformation matrix of reference frame $\Sigma_{m}$ with respect to frame $\Sigma_{fb}$, $\bld{G}_{fb}^{m}$, is constant.

If we denote the floating-base joint velocity and acceleration vectors as $\dot{\bld{q}}_{\!fb}$ and $\ddot{\bld{q}}_{\!fb}$, respectively, its twist $\bld{\nu}_{\!fb}\in se(3)$ and the spatial acceleration $\dot{\bld{\nu}}_{\!fb}\in se(3)$ of each joint reference frame from $\Sigma_{w}$ to $\Sigma_{tcp}$ can be recursively evaluated along the serial kinematic structure of the floating base system:
\begin{eqnarray}
	\bld{\nu}_i & = & \mbox{Ad}_{G_i^{\lambda}} \bld{\nu}_{\lambda} + \bld{s}_i\dot{q}_i
	\label{eq:twist} \\
    \dot{\bld{\nu}}_i & = & \mbox{Ad}_{G_i^{\lambda}} \dot{\bld{\nu}}_{\lambda} + \bld{s}_i\ddot{q}_i + \mbox{ad}_{\nu_i}\bld{s}_i\dot{q}_i
	\label{eq:accel}
\end{eqnarray}
where $\lambda$ denotes the parent reference frame of frame $i$, $\mbox{ad}_{(\cdot)}$ is the Lie bracket operator, $\dot{q}_i$ and $\ddot{q}_i$ are the velocity and acceleration of the $i$-th joint, and $\mbox{Ad}_{G_i^{\lambda}}$ is the adjoint operator:
\begin{equation}
\mbox{Ad}_{G_i^{\lambda}} \ = \ \mbox{Ad}_{e^{[{\scriptsize \bld{s}}]q}} \ = \ e^{\mbox{ad}_{{\scriptsize \bld{s}} q}}
	\label{eq:Adj}
\end{equation}

At the first recursive iteration of (\ref{eq:twist}-\ref{eq:accel}), consider the root conditions at frame $\Sigma_{w}$, i.e., $\bld{\nu}_{\lambda} = \bld{\nu}_{w} = \bld{0}$ and $\dot{\bld{\nu}}_{\lambda} = \dot{\bld{\nu}}_{w} = \dot{\bld{\nu}}_{g}$, where $\dot{\bld{\nu}}_{g}$ is the spatial acceleration caused by gravity.

Let us now define the rigid-body spatial momentum as
\begin{eqnarray}
	\bld{\mu} \ = \ \M\bld{\nu} \ \in \ se^{*}(3)
	\label{eq:mu}
\end{eqnarray}
where $se^{*}(3)$ is the dual Lie algebra associated with $SE(3)$ and $\M\in\real^{6\times6}$ is the spatial inertia matrix defined as
\begin{eqnarray}
	\label{eq:spinertia}
	\bld{M} & = &
	\begin{bmatrix}
		m\bld{I}_3 & -m[\bld{r}^{\tiny\mbox{cm}}] \\
		m[\bld{r}^{\tiny\mbox{cm}}] & \bld{\mathcal{I}} -m[\bld{r}^{\tiny\mbox{cm}}]^2
	\end{bmatrix}
\end{eqnarray}
where $m\in \real$ is the mass of the body, $\bld{r}^{\tiny\mbox{cm}}\in\real^3$ is its center of mass expressed at local coordinates, $\bld{\mathcal{I}}\in\real^{3\times 3}$ is its inertia tensor, and $\bld{I}_3\in\real^{3\times 3}$ is an identity matrix.

\subsection{Parallel-Serial Manipulator Dynamics}
\label{sec:manip}
The configurational vector of the parallel-serial manipulator is denoted as $\bld{q}_{arm}\!\in\!\real^{n_r}$, and $\dot{\bld{q}}_{arm},\ddot{\bld{q}}_{arm}\!\in\!\real^{n_r}$ are its corresponding velocity and acceleration where $n_r$ is the manipulator's DoF. Thus, the second-order differential forward kinematics from frame $\Sigma_{w}$ to $\Sigma_{tcp}$ can be computed by recursively applying (\ref{eq:op00})-(\ref{eq:Adj}) to obtain the end-effector twist $\bld{\nu}_{\!tcp}$ and spatial accelerations $\dot{\bld{\nu}}_{\!tcp}$. If we denote the transformation matrix from $\Sigma_{m}$ to $\Sigma_{tcp}$ as $\bld{G}_{\!m}^{tcp}\!(\bld{q}_{arm})$, the forward kinematic from $\Sigma_{w}$ to $\Sigma_{tcp}$ is
\begin{equation}
	\bld{G}_{\!w}^{tcp} \ = \ \bld{G}_{\!w}^{fb} 
\, \bld{G}_{\!fb}^{m} \, \bld{G}_{\!m}^{tcp}
	\label{eq:fwdKin}
\end{equation}

By computing the spatial momentum (\ref{eq:mu}) of all rigid bodies composing the manipulator, the wrench $\F\in se^{*}(3)$ acting on each rigid body can be calculated from the controlled Euler-Poincar\'{e} equation \cite{marsden2013introduction}, as follows:
\begin{equation}
	\F \ = \ \M\dot{\bld{\nu}} - \mbox{ad}_{\nu}^*\left(\bld{\mu} \right)
	\label{eq:wrench}
\end{equation}
where $\mbox{ad}^{*}_{(\cdot)}$ is the adjoint operator for the dual Lie algebra. See \cite{paz2024analytical} for further details on adjoint operators.

The analytic dynamic of parallel-serial manipulators is obtained by solving the forces in the actuator's joints \cite{petrovic2022mathematical}\cite{alvaro2024analytical}. Moreover, the manipulator's total wrench $\F_{\!m}$ acting on the reference frame $\Sigma_{m}$ can be computed by projecting the wrenches of all rigid bodies composing the manipulator:
\begin{equation}
	\F_{\!m} \ = \ \sum_i \mbox{Ad}_{G_{i}^{m}}^{*} \F_{\!i}
	\label{eq:sumForce}
\end{equation}
where $\mbox{Ad}_{G}^*$ is the adjoint operator for elements in $se^*(3)$ and $\F_{\!i}$ is the wrench of the $i$-th rigid body in local coordinates.
\subsection{Actively-Articulated Suspension Mechanism}
\begin{figure}[h!] 
	\centering
	\includegraphics[trim={0cm 0cm 0cm 0.16cm},clip,width=8cm]{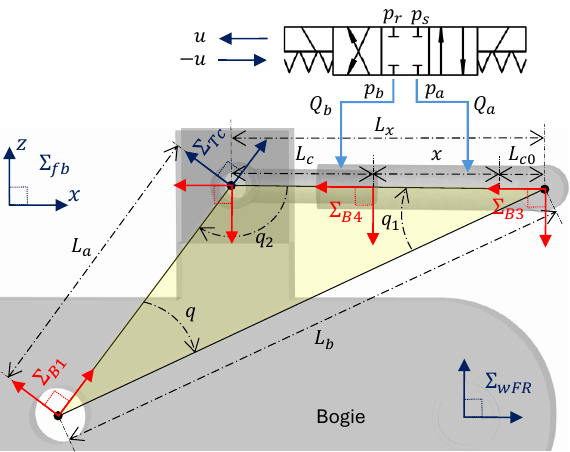}
	\vspace*{0cm}
	\caption{{\footnotesize {\bf Geometry of the articulated suspension with linear actuator.} This one-DOF closed kinematic chain can be locally analyzed in $SE(2)$, i.e. plane $x \!\cdot\! z$, and it is articulated by three passive rotational joints placed at reference frames $\Sigma_{\!B1}$, $\Sigma_{\!B3}$ and $\Sigma_{\!Tc}$, and a linearly actuated one at $\Sigma_{\!B4}$ which is powered by a hydraulic valve system.}}
	\label{fig:suspension}
\end{figure}

The base rigid body and the chassis are connected through two suspension mechanisms, as depicted in Figs. \ref{fig:graph} and \ref{fig:snaps}. The right-side suspension mechanism is shown in Fig. \ref{fig:suspension}, and the left-side suspension is assumed to be symmetric; thus, its mechanical analysis is similar. The suspension structure can be analyzed as a parallel mechanism \cite{alvaro2024analytical} or a four-joint closed kinematic chain. The three passive rotational joints located at reference frames $\Sigma_{B1}$, $\Sigma_{B3}$, and $\Sigma_{Tc}$, with angular positions $\theta$, $\theta_1$, and $\theta_2$, respectively, generate a triangle with lengths $L_{a}$, $L_{b}$, and $L_{x} > 0$ (Fig. \ref{fig:suspension}). Distances $L_{a}$ and $L_{b}$ are constant and $L_{x} = L_{c0} + x + L_{c}$, where $x$ is the actuator's position and $L_{c0}, L_{c} \geq 0$ are also constant. Kinematically, this mechanism can be studied as two serial kinematic chains with the following reference frame sequences:
\begin{eqnarray}
	&\Sigma_{fb} \ \rightarrow \ \Sigma_{B3} \ \rightarrow \ \Sigma_{B4} \ \rightarrow \ \Sigma_{Tc} \label{eq:seqs} \\
    &\Sigma_{fb} \ \rightarrow \ \Sigma_{B1} \ \rightarrow \ \Sigma_{Tc} \nonumber
\end{eqnarray}
such that both chains are restricted by holonomic constraints by imposing the same initial and final reference frames.

The trigonometry is resolved by the inner angles $q , q_1 , q_2 \!<\! 0$, which are functions of $x$, by using the expressions \cite{zhu2010virtual}
	\begin{eqnarray}
		q & = & -\!\arccos\!\left(\! \tfrac{L_{x}^{2}-L_{b}^{2}-L_{a}^{2}}{-2 L_{b} L_{a} } \right) \label{eq:qus_a} \\
		q_{1} & = & -\!\arccos\!\left(\! \tfrac{L_{a}^{2}-L_{x}^{2}-L_{b}^{2}}{-2L_{x}L_{b}} \right) \label{eq:qus_b} \\
		q_{2} & = & -\!\arccos\!\left(\! \tfrac{L_{b}^{2}-L_{x}^{2} - L_{a}^{2}}{-2L_{x}L_{a}} \right) \label{eq:qus_c}
	\end{eqnarray}

In addition, the angles $\theta, \theta_1, \theta_2$ and $q, q_1, q_2$ rotate around the same axes of motion of $\Sigma_{B1}$, $\Sigma_{B3}$, and $\Sigma_{Tc}$, respectively. Thus, the passive joint angles are related by linear forms as
\begin{equation}
	\theta \ = \ q + \psi, \quad \theta_1 \ = \ q_1 + \psi_1, \quad \theta_2 \ = \ q_2 + \psi_2
 \label{eq:thetas}
\end{equation}
where $\psi$, $\psi_1$, and $\psi_2$ are constant offset angles. In addition, by time differentiating the last expressions, we obtain linear forms of the passive joint velocities and accelerations:
\begin{eqnarray}
	&\hspace{-1.5cm} \dot{\theta} \ \ = \ \dot{q} \ \ = \ k_1 \dot{x}, \qquad & \ddot{\theta} \ \ = \ \ddot{q} \ \ = \ \dot{k}_1 \dot{x} + k_1 \ddot{x} \\
    &\hspace{-1.5cm} \dot{\theta}_{1} \ = \ \dot{q}_{1} \ = \ k_2 \dot{x}, \qquad & \ddot{\theta}_1 \ = \ \ddot{q}_{1} \ = \ \dot{k}_2 \dot{x} + k_2 \ddot{x} \\
    &\hspace{-1.5cm} \dot{\theta}_{2} \ = \ \dot{q}_{2} \ = \ k_3 \dot{x}, \qquad & \ddot{\theta}_2 \ = \ \ddot{q}_{2} \ = \ \dot{k}_3 \dot{x} + k_3 \ddot{x}
	\label{eq:dq__}
\end{eqnarray}
where $\dot{x}$ and $\ddot{x}$ are the linear velocity and acceleration of the actuator, and $\dot{k}_1, \dot{k}_2, \dot{k}_3$ are the time derivatives of $k_1, k_2, k_3$, which can be computed 
following the method adopted in \cite{alvaro2024analytical}.

Equation (\ref{eq:seqs}) assumes that both kinematic chains have the same initial and final reference frames. However, the geometric closure of the closed kinematic chain is forced through  (\ref{eq:qus_a}-\ref{eq:qus_c}) since $q \!+\! q_1 \!+\! q_2 \!+\!\pi \!=\! 0$, and the holonomic constraints are forced kinematically via the same initial and final absolute transformation matrices $\bld{G}_{\!fb}, \bld{G}_{\!Tc}$, twists $\bld{\nu}_{\!fb}, \bld{\nu}_{_{\!Tc}}$, and spatial accelerations $\dot{\bld{\nu}}_{\!fb}, \dot{\bld{\nu}}_{_{\!Tc}}$ in both chains. Once the joint state and the floating base twist $\bld{\nu}_{\!fb}$ and spatial acceleration $\dot{\bld{\nu}}_{\!fb}$ are computed, we can use (\ref{eq:twist}) and (\ref{eq:accel}) for computing the twists at each reference frame of the mechanism as
	\begin{subequations}
		\begin{eqnarray}
			\bld{\nu}_{\!_{\!B1}} & = & \mbox{Ad}_{G_{\!_{B1}}^{_{fb}}}\bld{\nu}_{\!_{\!fb}} + \bld{s}_{z}\dot{q} \label{eq:nu_b1} \\
			\bld{\nu}_{\!_{\!Tc}} & = & \mbox{Ad}_{G_{\!_{Tc}}^{\!_{B1}}}\bld{\nu}_{\!_{\!B1}} \label{eq:nu_e} \\
			\bld{\nu}_{\!_{\!B3}} & = & \mbox{Ad}_{G_{\!_{B3}}^{_{fb}}}\bld{\nu}_{\!_{\!fb}} + \bld{s}_{z}\dot{q}_{1} \label{eq:nu_b3} \\
			\bld{\nu}_{\!_{\!B4}} & = & \mbox{Ad}_{G_{\!_{B4}}^{\!_{B3}}}\bld{\nu}_{\!_{\!B3}} + \bld{s}_{x}\dot{x} \label{eq:nu_b4} \\
			\bld{\nu}_{\!_{\!Tc}} & = & \mbox{Ad}_{G_{\!_{Tc}}^{\!_{B4}}}\bld{\nu}_{\!_{\!B4}} + \bld{s}_{z}\dot{q}_{2} \label{eq:nu_t2}
		\end{eqnarray}
		\label{eq:36}
	\end{subequations}
and spatial accelerations as
	\begin{subequations}
		\begin{eqnarray}
			\dot{\bld{\nu}}_{\!_{\!B1}} & = & \mbox{Ad}_{G_{\!_{B1}}^{_{fb}}}\dot{\bld{\nu}}_{\!_{\!fb}} + \bld{s}_{z}\ddot{q} + \mbox{ad}_{\nu_{B1}}\bld{s}_z\dot{q} \label{eq:dnu_b1} \\
			\dot{\bld{\nu}}_{\!_{\!Tc}} & = & \mbox{Ad}_{G_{\!_{Tc}}^{\!_{B1}}}\dot{\bld{\nu}}_{\!_{\!B1}} \label{eq:dnu_e} \\
			\dot{\bld{\nu}}_{\!_{\!B3}} & = & \mbox{Ad}_{G_{\!_{B3}}^{_{fb}}}\dot{\bld{\nu}}_{\!_{\!fb}} + \bld{s}_{z}\ddot{q}_{1} + \mbox{ad}_{\nu_{B3}}\bld{s}_z\dot{q}_1 \label{eq:dnu_b3} \\
			\dot{\bld{\nu}}_{\!_{\!B4}} & = & \mbox{Ad}_{G_{\!_{B4}}^{\!_{B3}}}\dot{\bld{\nu}}_{\!_{\!B3}} + \bld{s}_{x}\ddot{x} + \mbox{ad}_{\nu_{B4}}\bld{s}_x\dot{q} \label{eq:dnu_b4} \\
			\dot{\bld{\nu}}_{\!_{\!Tc}} & = & \mbox{Ad}_{G_{\!_{Tc}}^{\!_{B4}}}\dot{\bld{\nu}}_{\!_{\!B4}} + \bld{s}_{z}\ddot{q}_{2} + \mbox{ad}_{\nu_{Tc}}\bld{s}_z\dot{q}_2 \label{eq:dnu_t2}
		\end{eqnarray}
		\label{eq:53}
	\end{subequations}
	where the screw axes of motion are
	\begin{eqnarray*}
		\bld{s}_{x} & = & {\footnotesize \begin{bmatrix}
			1 & 0 & 0 & 0 & 0 & 0
		\end{bmatrix}}^{\top} \\
		\bld{s}_{z} & = & {\footnotesize \begin{bmatrix}
			0 & 0 & 0 & 0 & 0 & 1
		\end{bmatrix}}^{\top}
	\end{eqnarray*}

The aforementioned analysis can be extended to the left closed kinematic chain, thus the general suspension state is $\bld{x}\!=\![x_R, x_L]^\top \!\in\! \real^{2}$ which contains the positions of right and left actuators and being $\dot{\bld{x}}$ the velocity and $\ddot{\bld{x}}$ the acceleration.
\subsection{Analytic Multibody Inertial Parameters}
The mobile platform depicted in Fig. \ref{fig:snaps} is supported by four wheels on the ground. Such wheels have no steering actuation, which means the platform steering is done by wheels velocity differential. Therefore, each wheel can be mechanically modeled as a one-DoF rotational mechanism and, from Fig. \ref{fig:suspension}, we know that the frames $\Sigma_{\!B1}$ and wheel frame $\Sigma_{wFR}$ are attached to the bogie rigid body resulting into a constant kinematic transformation among them. Now we thus define the four wheels angular position as
	\begin{equation}
		\bld{q}_{w} \ = \ \begin{bmatrix}
			q_{wFR} & q_{wFL} & q_{wRR} & q_{wRL}
		\end{bmatrix}^{\top}
	\end{equation}
where sub indexes refer to the FrontRear/RightLeft wheel. Then the wheels angular velocities and accelerations are denoted by $\dot{\bld{q}}_w$ and $\ddot{\bld{q}}_w$ which into expressions (\ref{eq:op00}-\ref{eq:accel}) generate the second order forward kinematics up to the wheels reference frames assuming that $\bld{G}_{w}^{_{B1}}$ is known.

The state of the whole platform is parameterized by the vectors $\bld{q}_{\!fb}$, $\bld{q}_{arm}$, $\bld{x}$, and $\bld{q}_{w}$. Figs. \ref{fig:platform}, \ref{fig:snaps} and \ref{fig:suspension} show that, discarding the manipulator, the mobile platform is composed of eleven rigid bodies, the base spatial inertia $\bld{M}_{\!fb}$ attached to frame $\Sigma_{fb}$, and five spatial inertias, on the right and left sides, attached to the following reference frames: bogie inertia is attached to $\Sigma_{B1}$, actuator cylinder and piston to $\Sigma_{B3}$ and $\Sigma_{B4}$, and front and rear wheels to $\Sigma_{wF}$ and $\Sigma_{wR}$, respectively. According to Fig. \ref{fig:snaps}, the chassis frame $\Sigma_{c}$ is located at the center of both bogies and aligned with the terrain. Then, all spatial inertias of both the manipulator and mobile platform can be expressed at $\Sigma_c$ by using the articulated-body inertia method \cite{Featherstone1987}, as follows:
\begin{equation}
\bld{M}_{c} \ = \ \sum_i \mbox{Ad}^*_{{G_{i}^{c}}} \bld{M}_i \mbox{Ad}_{{G_{i}^{c}}}
\label{eq:84c}
\end{equation}
where the index $i$ iterates all over the reference frames $\Sigma_i$ containing an attached rigid-body spatial inertia matrix $\bld{M}_i$, and $\bld{G}_{i}^{c}\in SE(3)$ is the transformation matrix from $\Sigma_i$ to $\Sigma_c$. The spatial inertia matrix $\bld{M}_{c}$ accumulates all mobile-manipulator inertias in a single frame and is parameterized by the joint positions. Additionally, it can be expressed in the world reference frame by
\begin{equation}
\bld{M}_{w} \ = \ \mbox{Ad}^*_{{G_{c}^{w}}} \bld{M}_c \mbox{Ad}_{{G_{c}^{w}}}
\label{eq:845}
\end{equation}

By following the inner composition of spatial inertias in (\ref{eq:spinertia}), the mobile-manipulator mass $m_c$, the center of mass $\bld{r}^{\tiny\mbox{cm}}_c$, and the inertia tensor $\bld{\mathcal{I}}_c$ can be extracted from $\bld{M}_c$ as
\begin{eqnarray}
    m_c & = & \bld{M}_{\!c\,(1,1)} \\
    \bld{r}^{\tiny\mbox{cm}}_c & = & m_c^{-1}\begin{bmatrix}
		\bld{M}_{\!c\,(6,2)} & \bld{M}_{\!c\,(4,3)} & \bld{M}_{\!c\,(5,1)}
	\end{bmatrix}^{\top} \\
    \bld{\mathcal{I}}_c & = & \bld{M}_{\!c\,(4:6,4:6)} + m_{c}[\bld{r}^{\tiny\mbox{cm}}_{c}]^2
\end{eqnarray}
where the sub-indexes $(\cdot,\cdot)$ follow a M{\scriptsize ATLAB} notation, which means that they refer to elements of the matrix $\bld{M}_{\!c}$. Let us assume that the mobile-platform center-of-mass reference frame $\Sigma_{\tiny\mbox{cm}}$ is located at $\bld{r}^{\tiny\mbox{cm}}_c$ and aligned with the chassis frame $\Sigma_c$. Then, $\bld{G}_{\!w}^{\tiny\mbox{cm}}$ is obtained as $\bld{G}_{w}^{\tiny\mbox{cm}} = \bld{G}_{\!w}^{c}\bld{G}_{\!c}^{\tiny\mbox{cm}}$, where $\bld{G}_{\!c}^{\tiny\mbox{cm}}$ is parameterized by the pair $(\bld{r}^{\tiny\mbox{cm}}_c, \bld{I}_3)\in SE(3)$ where $\bld{I}_3$ is the identity matrix since $\bld{R}^{\tiny\mbox{cm}}_c = \bld{I}_3 \in SO(3)$. Moreover, the inertia tensor expressed at frame $\Sigma_{\tiny\mbox{cm}}$ is $\bld{\mathcal{I}}_{\tiny\mbox{cm}} = \bld{R}^{\tiny\mbox{cm}}_c \bld{\mathcal{I}}_c \bld{R}^{\tiny\mbox{cm}\top}_c  = \bld{\mathcal{I}}_c$.
\subsection{Inertial-Variant Wheel Normal Forces}
\label{sec:forces}
We adopt the analytic wheel normal force calculation reported in \cite{petrovic2022analytic}, where the four wheels' normal forces $f_{\!F\!R}^n$, $f_{\!F\!L}^n$, $f_{\!R\!R}^n$, and $f_{\!R\!L}^n$ $\in \real$ are scalar functions of
\begin{itemize}
    \item The total wrench exerted by the manipulator $\F_{\!m}$ expressed at reference frame $\Sigma_{m}$. See equation (\ref{eq:sumForce}).
    \item The transformation matrix $\bld{G}_{\tiny\mbox{cm}}^{m}$ from reference frame $\Sigma_{\tiny\mbox{cm}}$ to the manipulator frame $\Sigma_{m}$.
    \item The transformation matrix $\bld{G}_{w}^{\tiny\mbox{cm}}$  from world frame $\Sigma_{w}$ to the mobile-platform center-of-mass frame $\Sigma_{\tiny\mbox{cm}}$.
    \item The twist vector $\bld{\nu}_{\tiny\mbox{cm}}$ and spatial acceleration $\dot{\bld{\nu}}_{\tiny\mbox{cm}}$ of frame $\Sigma_{\tiny\mbox{cm}}$ expressed at local coordinates.
    \item Distances of wheel frames $\Sigma_{wFR}$, $\Sigma_{wFL}$, $\Sigma_{wRR}$, and $\Sigma_{wRL}$ with respect to $\Sigma_{\tiny\mbox{cm}}$. Only in the axis $x$ of $\Sigma_{\tiny\mbox{cm}}$. These distances are generalized as $l_1$, $l_2$ in \cite{petrovic2022analytic} for rear and front wheels, respectively.
    \item Mobile-platform inertia tensor $\bld{\mathcal{I}}_c$.
\end{itemize}

The formulation reported in \cite{petrovic2022analytic} assumes the mobile platform as a single rigid body where inertial parameters are constant. By including an articulated suspension between the base rigid body and the chassis bogies, the mobile platform becomes an articulated multibody system by itself the configuration of which is given by the suspension actuators' position $\bld{x}$. Therefore, the multibody center of mass and inertia tensor become direct functions of $\bld{x}$ and, consequently, all aforementioned enlisted elements. Even the manipulator's wrench $\F_{\!m}$ becomes a function of $\bld{x}$ since, before performing the iterative-recursive dynamic algorithm described in \ref{sec:manip}, the matrix $\bld{G}_{\!w}^{m}$, twist $\bld{\nu}_{\!m}$, and spatial accelerations $\dot{\bld{\nu}}_{\!m}$ of $\Sigma_m$ must be computed, which are functions of $\bld{x}$, $\dot{\bld{x}}$, and $\ddot{\bld{x}}$.
\subsection{Hydraulics of Suspension Actuators}
\label{sec:hydra}

Hydraulic valves for controling the supension actuators are depicted in Fig. \ref{fig:suspension}, where $p_a$ and $p_b$ are the cylinder chambers pressures, $p_s$ is the system supply pressure, $p_r$ is the tank pressure and $Q_a$ and $Q_b$ are the flow rates of the cylinder chambers which can be computed as \cite{koivumaki2015stability}
\begin{eqnarray}
    Q_a & = & c_{p1}v(p_s - p_a)uS(u) + c_{n1}v(p_a - p_r)uS(-u) \nonumber \\
    Q_b & = & -c_{n2}v(p_b - p_r)uS(u) + c_{p2}v(p_s - p_b)uS(-u) \nonumber
\end{eqnarray}
where $c_{p1}$, $c_{n1}$, $c_{p2}$, $c_{n2}>0$ are flow coefficients of the valve control $u$, the pressure drop function is $v(\Delta p) = \mbox{sign}(\Delta p)\sqrt{|\Delta p|}$ and $S(u)$ is the selector function defined by
\begin{equation}
    S(u) = \biggl\{ \begin{matrix}
        1, & \mbox{if} \quad u>0 \\
        0, & \mbox{if} \quad u \leq 0
    \end{matrix}
\end{equation}

The time derivatives of the pressures inside the cylinder chambers are retrieved through
\begin{eqnarray}
    \dot{p}_a & = & \dfrac{\beta}{A_a x}(Q_a - A_a \dot{x}) \\
    \dot{p}_b & = & \dfrac{\beta}{A_b (s - x)}(Q_b + A_b \dot{x})
\end{eqnarray}
where $\beta$ stands for the bulk modulus of the fluid, $s$ is the maximum stroke of the cylinder piston and $A_a$ and $A_b$ are the piston areas inside the chambers. Thus, the linear force $f_p$ delivered by the hydraulic piston is given by
\begin{equation}
    f_p \ = \ A_a p_a - A_b p_b
\end{equation}
\section{Nonlinear Optimal Control for Stability}
\label{sec:opt}
\begin{figure*}[h!]
	\centering
	\begin{minipage}[b]{0.32\textwidth}
		\includegraphics[trim={0.0cm 0.2cm 0.1cm 0.2cm},clip,width=\textwidth]{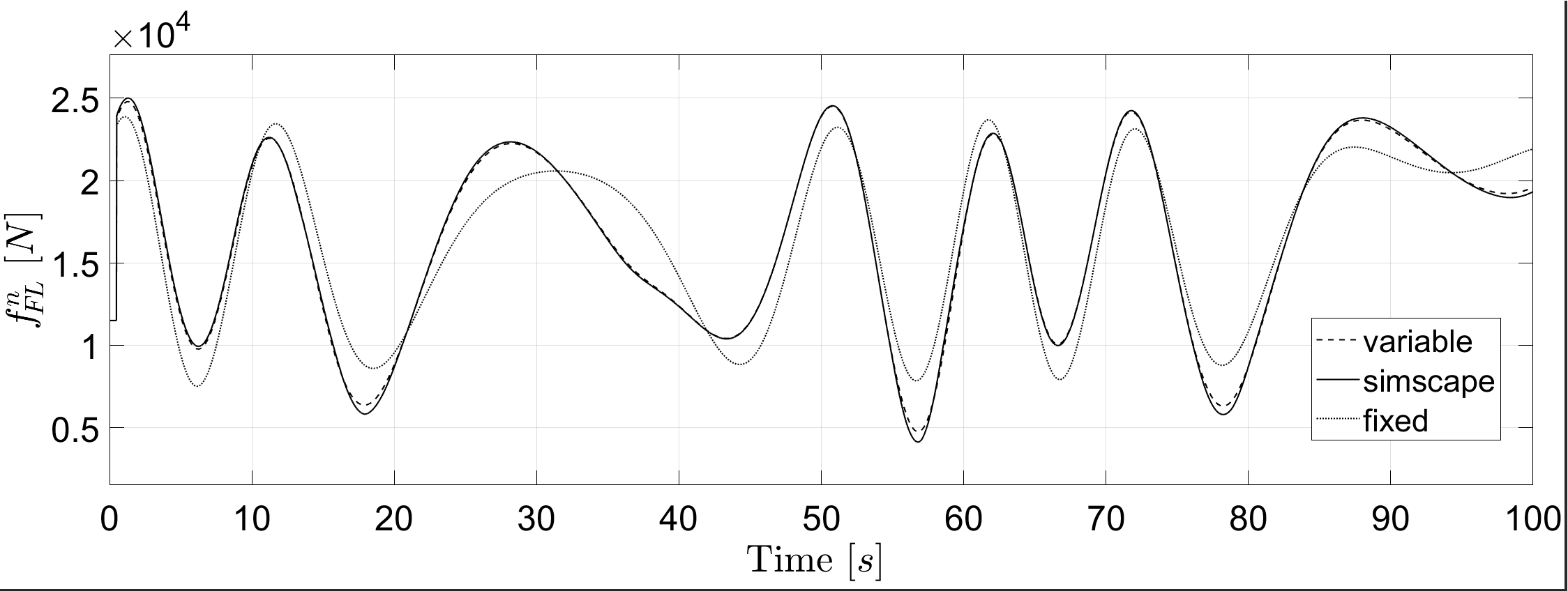}
	\end{minipage}
    \begin{minipage}[b]{0.32\textwidth}
		\includegraphics[trim={0.0cm 0.2cm 0.1cm 0.2cm},clip,width=\textwidth]{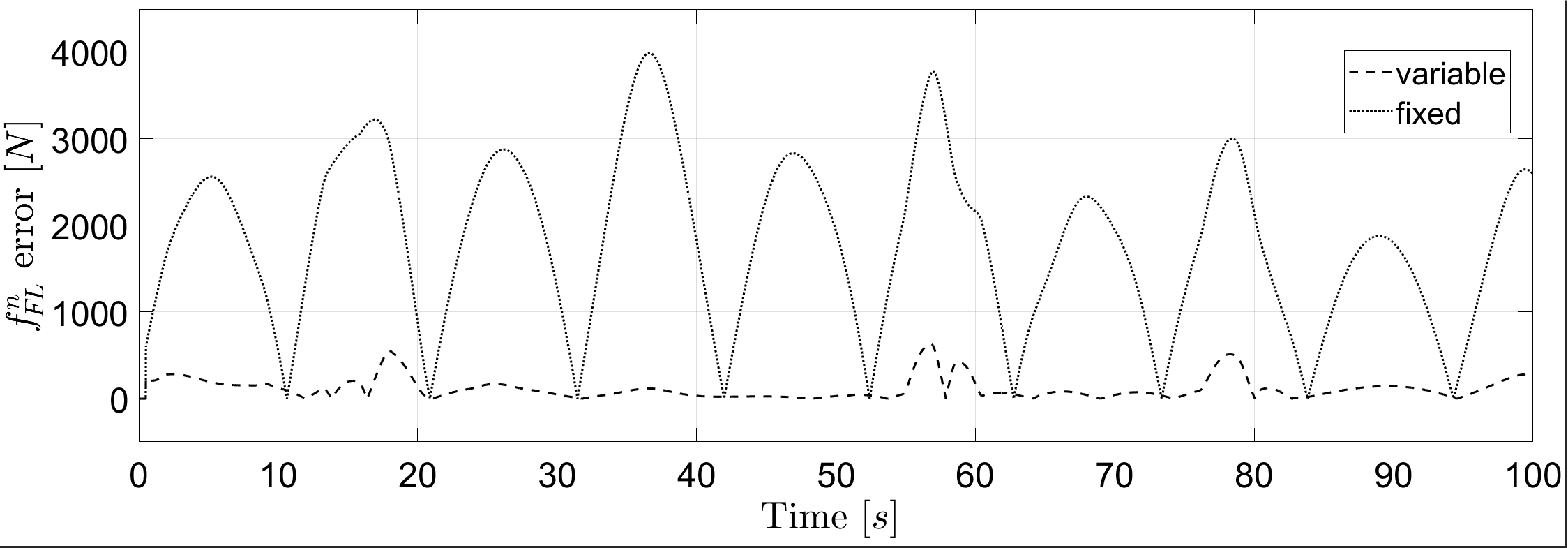}
	\end{minipage}
    \begin{minipage}[b]{0.32\textwidth}
		\includegraphics[trim={0.0cm 0.06cm 0.1cm 0.0cm},clip,width=\textwidth]{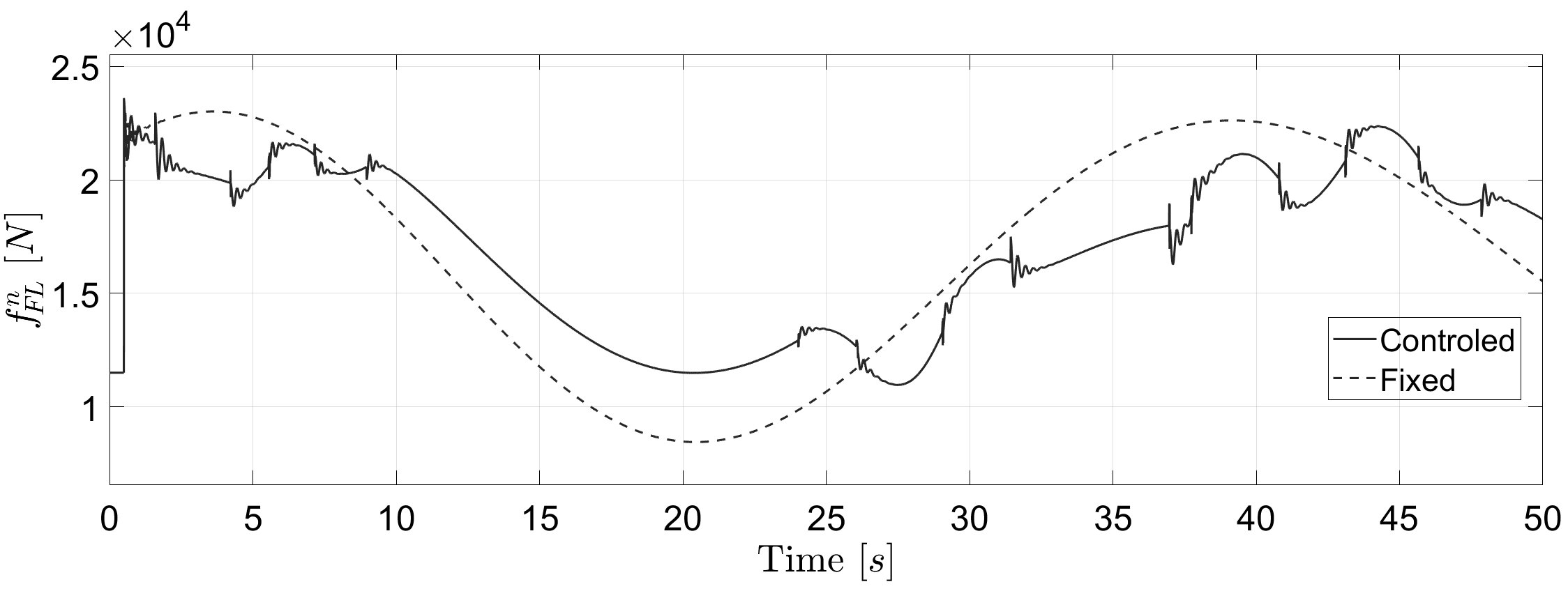}
	\end{minipage}    

    \vspace*{-0.0cm}
	\begin{minipage}[b]{0.32\textwidth}
		\includegraphics[trim={0.0cm 0.2cm 0.1cm 0.2cm},clip,width=\textwidth]{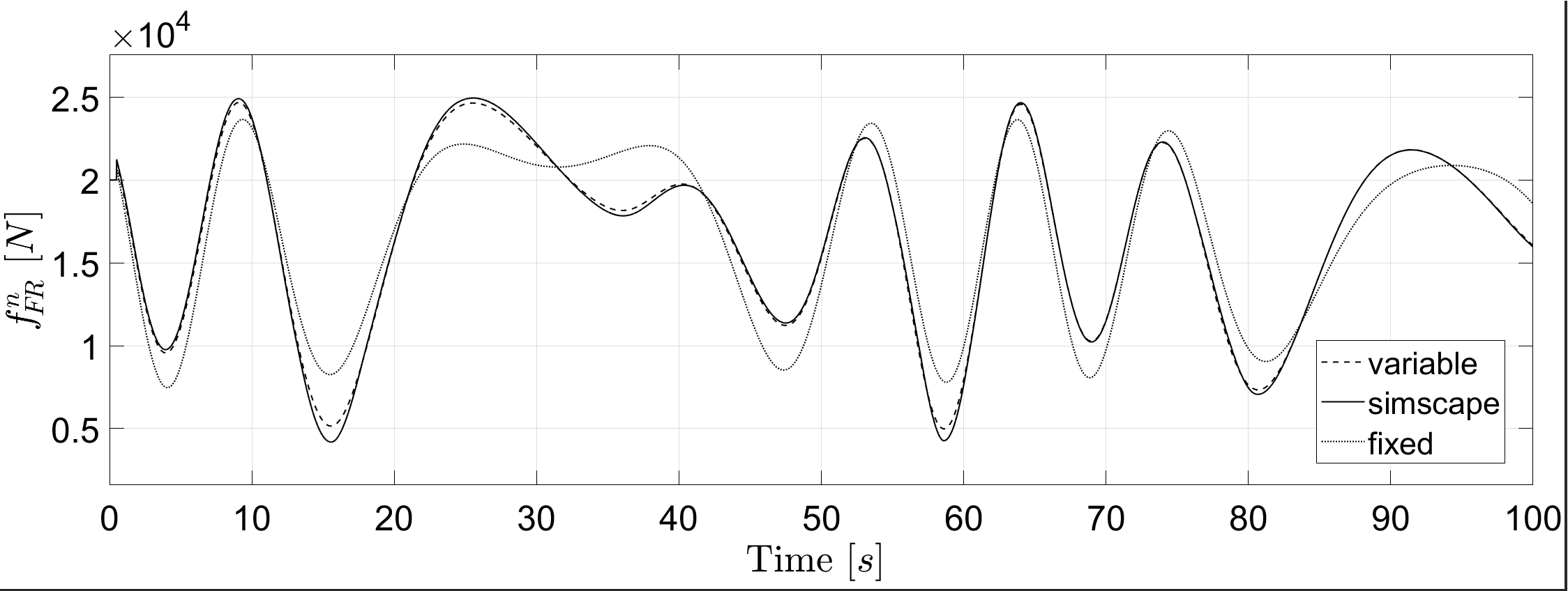}
	\end{minipage}
	\begin{minipage}[b]{0.32\textwidth}
		\includegraphics[trim={0.0cm 0.2cm 0.1cm 0.2cm},clip,width=\textwidth]{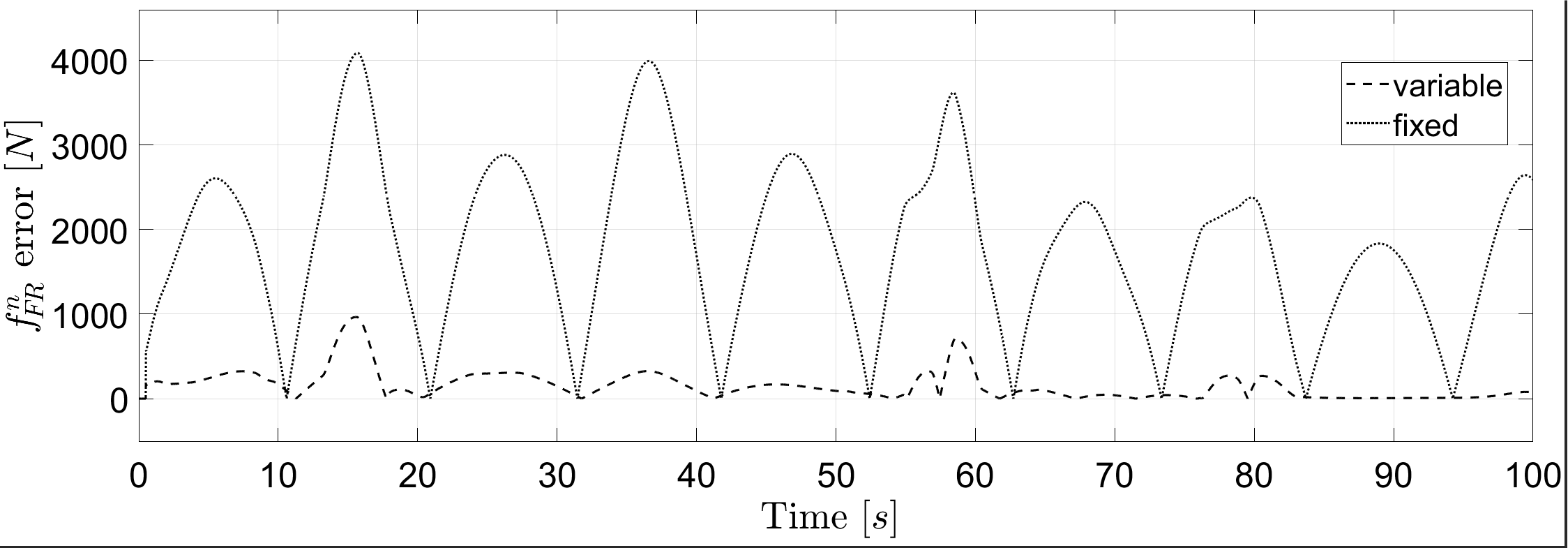}
	\end{minipage}
    \begin{minipage}[b]{0.32\textwidth}
		\includegraphics[trim={0.0cm 0.06cm 0.1cm 0.0cm},clip,width=\textwidth]{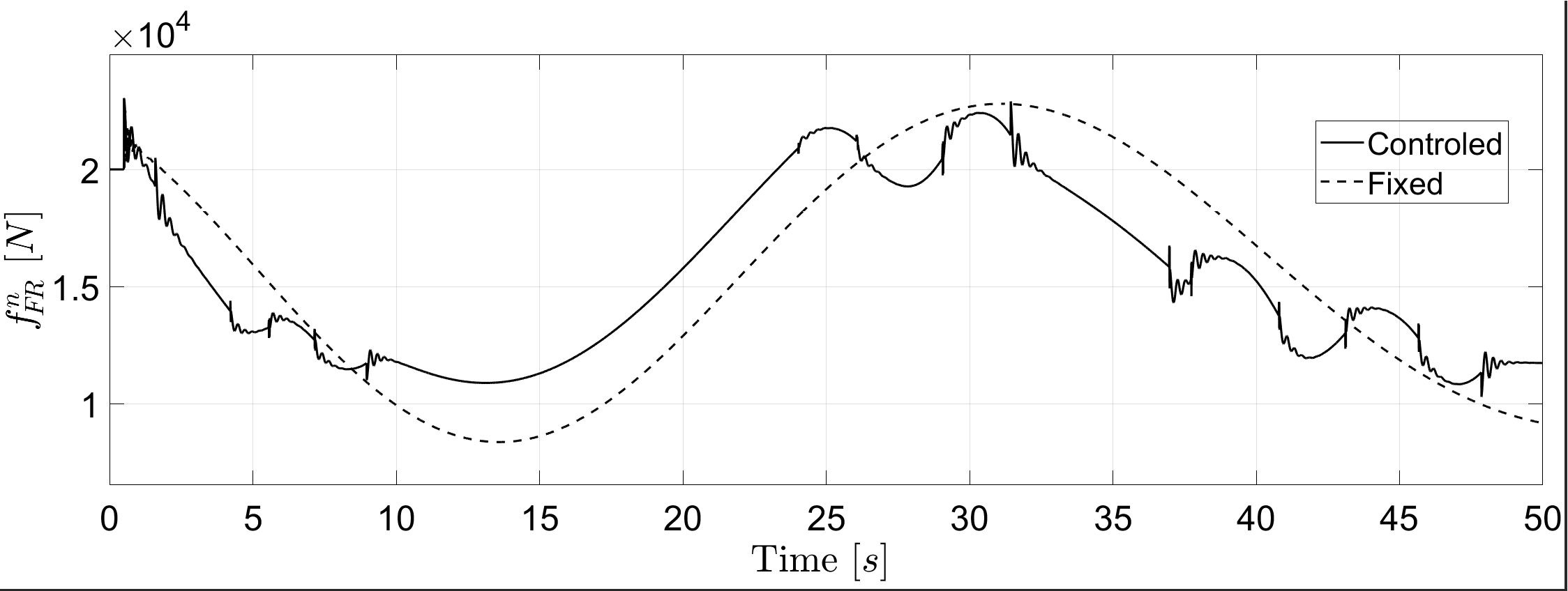}
	\end{minipage}

    \vspace*{-0.0cm}
	\begin{minipage}[b]{0.32\textwidth}
		\includegraphics[trim={0.0cm 0.2cm 0.1cm 0.2cm},clip,width=\textwidth]{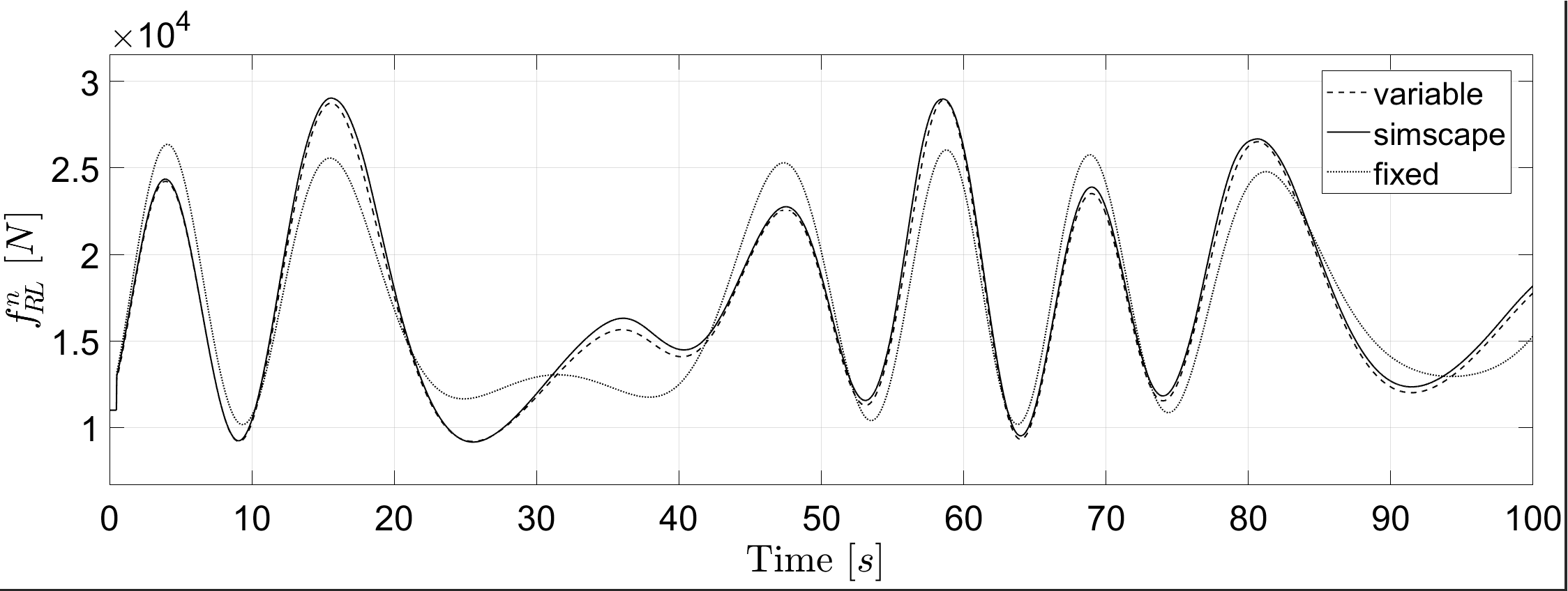}
	\end{minipage}
	\begin{minipage}[b]{0.32\textwidth}
		\includegraphics[trim={0.0cm 0.2cm 0.1cm 0.2cm},clip,width=\textwidth]{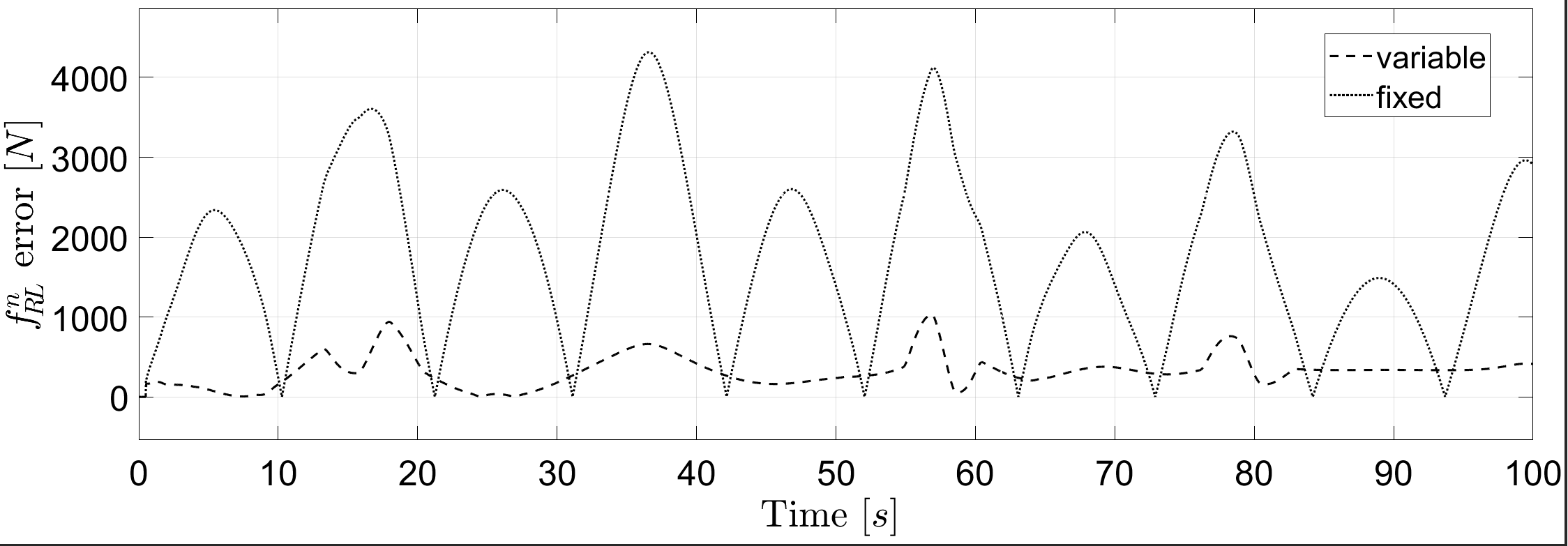}
	\end{minipage}
    \begin{minipage}[b]{0.32\textwidth}
		\includegraphics[trim={0.0cm 0.06cm 0.1cm 0.0cm},clip,width=\textwidth]{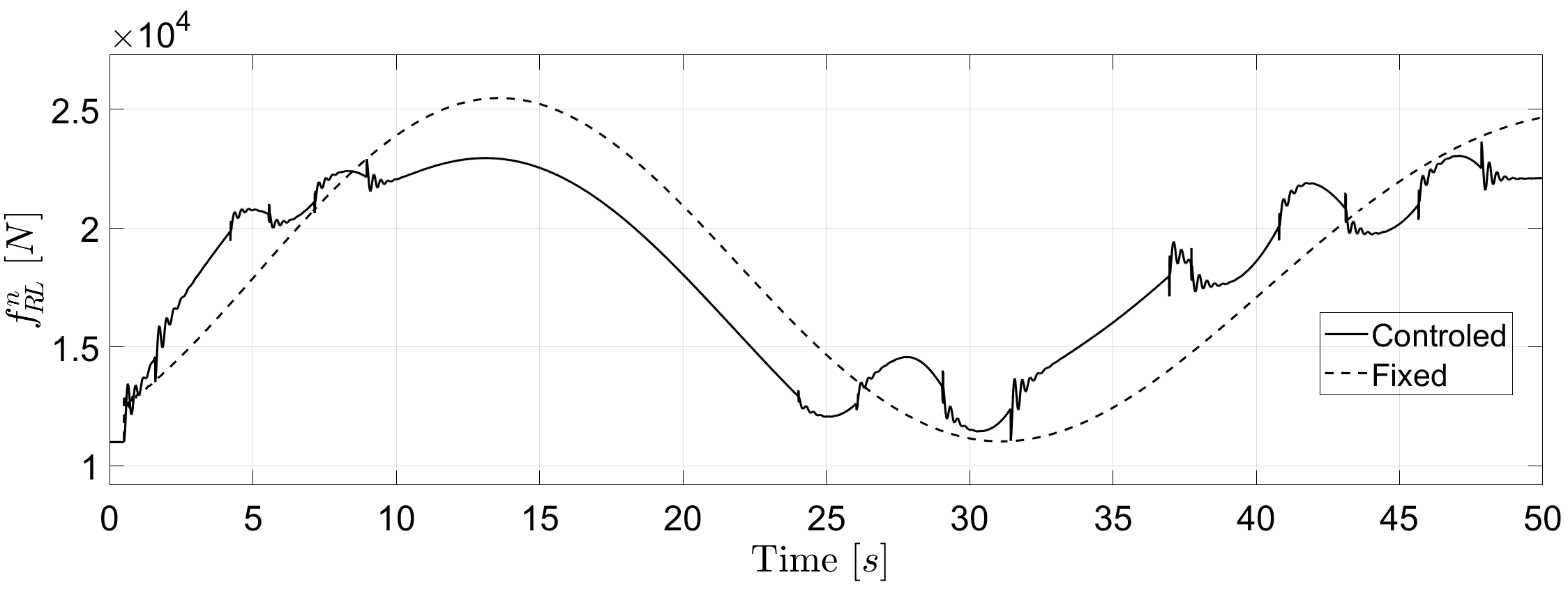}
	\end{minipage}    

    \vspace*{-0.0cm}
	\begin{minipage}[b]{0.32\textwidth}
		\includegraphics[trim={0.0cm 0.2cm 0.1cm 0.2cm},clip,width=\textwidth]{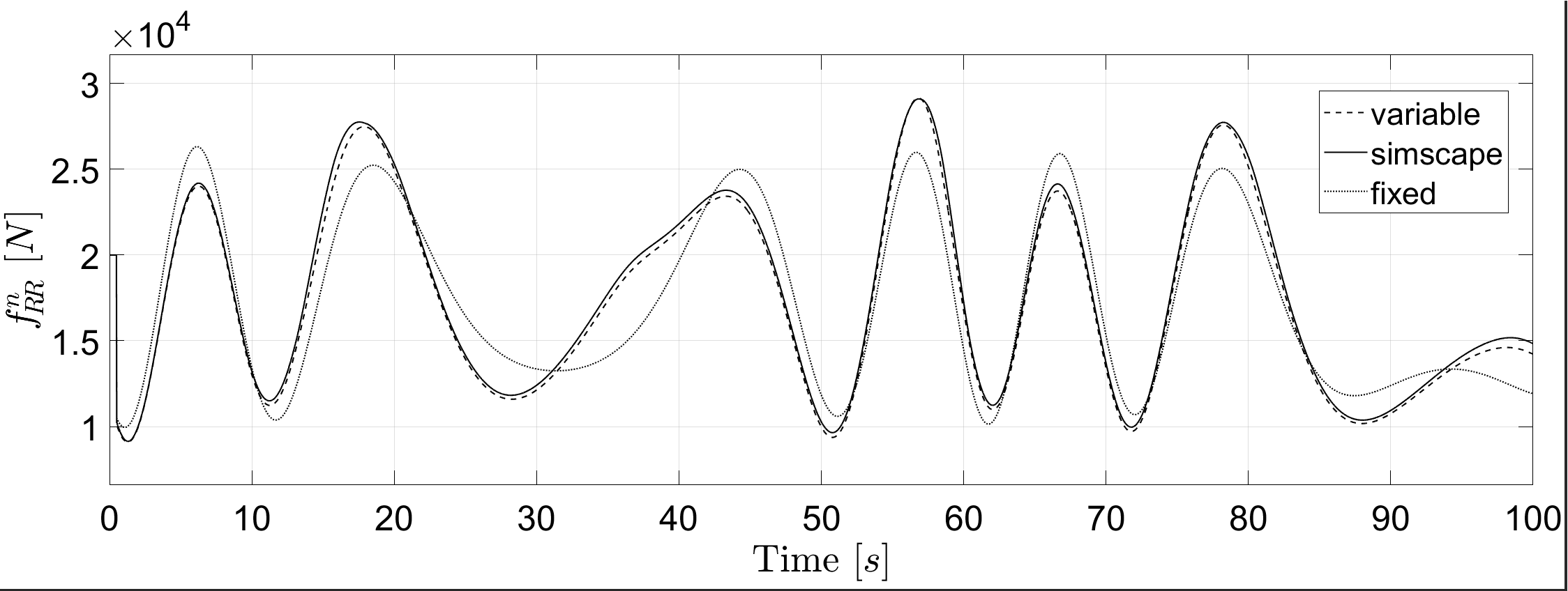}
        \caption{{\footnotesize {\bf Wheels normal forces.} Simscape is used as ground truth while variable and fixed inertial-parameters analytic solutions are depicted.}}
		\label{fig:forces_k}
	\end{minipage}
	\begin{minipage}[b]{0.32\textwidth}
		\includegraphics[trim={0.0cm 0.2cm 0.1cm 0.2cm},clip,width=\textwidth]{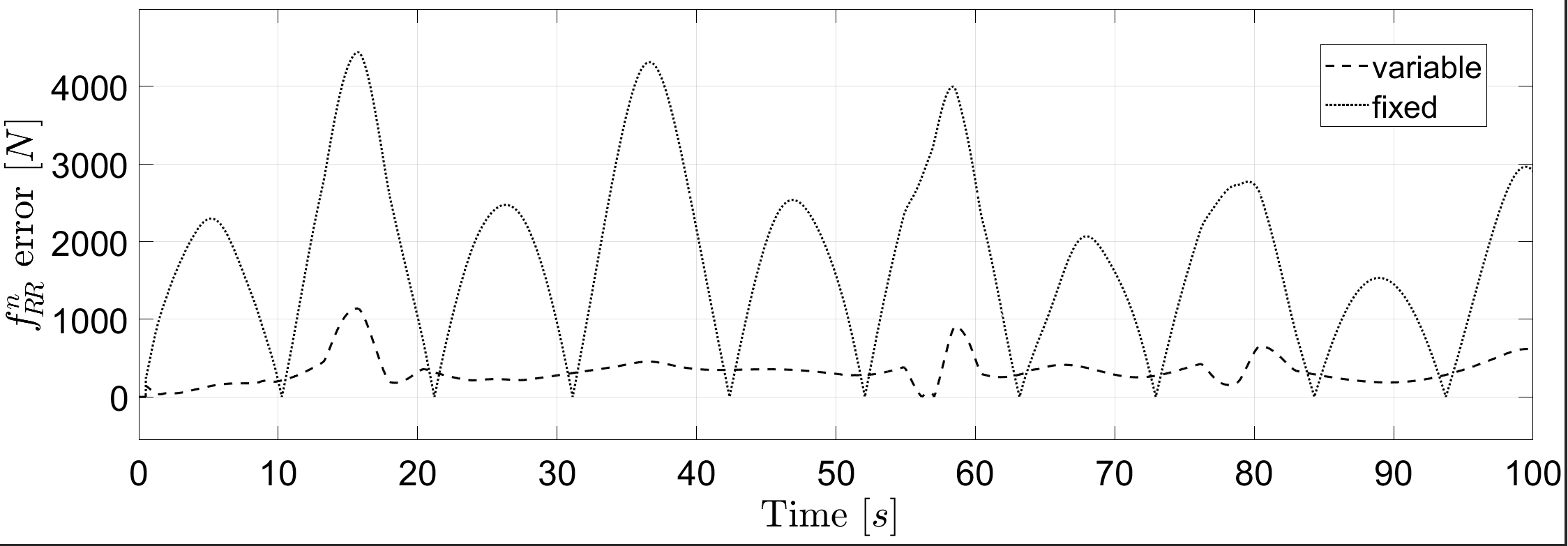}
        \caption{{\footnotesize {\bf Error in wheels normal forces.} Absolute error of variable and fixed analytic solutions with respect to the Simscape solution, see Fig. \ref{fig:forces_k}.}}
		\label{fig:forces_error}
	\end{minipage}
    \begin{minipage}[b]{0.32\textwidth}
		\includegraphics[trim={0.0cm 0.06cm 0.1cm 0.0cm},clip,width=\textwidth]{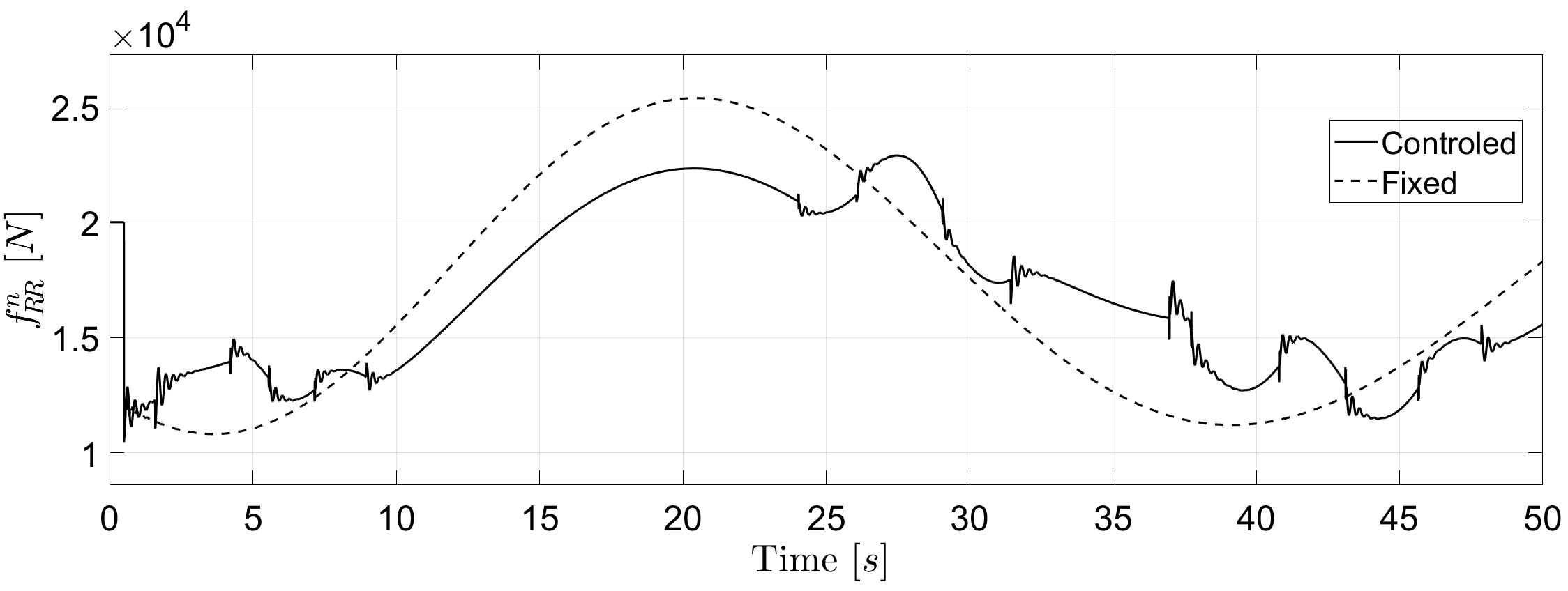}
        \caption{{\footnotesize {\bf Wheel normal forces with suspension control.} Normal forces of the controlled suspension behave better distributed than the fixed one.}}
		\label{fig:forces_control}
	\end{minipage}    
\end{figure*}
The static and dynamic stability of the mobile manipulator can be ensured through nonlinear optimal control, where at instant $k$, the solver has to find the next optimal suspension state $\bld{x}_{\!k\!+\!1}, \dot{\bld{x}}_{\!k\!+\!1}, \ddot{\bld{x}}_{\!k\!+\!1}$ as decision variables in (\ref{eq:costf2}). Moreover, the normal ground reaction forces described in \ref{sec:forces} can be used into the multiobjective cost function $f$ in (\ref{eq:costf2}), first term, where the difference among the four normal forces is penalized in order to maintain an optimal force distribution for dynamic stability. In contrast, the second term in (\ref{eq:costf2}) represents a static stability criterion where $\eta_\kappa$ indicates the stability angles reported in \cite{iagnemma2003control}, which are functions of $\bld{r}^{\tiny\mbox{cm}}_c$. Thus, assuming that the manipulator and wheel commands $\bld{q}_{arm}$ and $\bld{q}_{w}$ are functions of time, the next state of the suspension is retrieved from the constrained optimal control
\begin{equation}
	\hspace*{-0.1cm}\underset{\bld{x}_{\!k\!+\!1}, \dot{\bld{x}}_{\!k\!+\!1}, \ddot{\bld{x}}_{\!k\!+\!1}}{\operatorname{minimize}}  \ \ \ f \ = \quad \tfrac{1}{2} \hspace{-0.8cm} \sum_{\underset{i\neq j}{i,j\in\{ _{F\!R}, _{F\!L}, _{R\!R}, _{R\!L} \} } } \hspace{-0.8cm} \psi_i \left( f_{\!i}^n \!-\! f_{\!j}^n \right)^{2} + \sum_{\kappa=1}^{4} \dfrac{\psi_\kappa}{\eta_\kappa}
	\label{eq:costf2}
\end{equation}
\hspace{0.35cm}subject to
\begin{subequations}
\begin{eqnarray}
        \bld{0} &  \leq & f_{i}^n \qquad \forall \ i \ \in\{ _{F\!R}, _{F\!L}, _{R\!R}, _{R\!L}\! \} \label{const_a} \\[-2pt]
		\bld{x}_{min} &  \leq & \bld{x}_{k+1} \ \ \leq \ \ \bld{x}_{max} \label{const_b} \\[-2pt]
		\dot{\bld{x}}_{min}(\bld{x}_{k}) &  \leq & \dot{\bld{x}}_{k+1} \ \ \leq \ \ \dot{\bld{x}}_{max}(\bld{x}_{k}) \label{const_c} \\[-2pt]
		\ddot{\bld{x}}_{min}(\bld{x}_{k}) &  \leq & \ddot{\bld{x}}_{k+1} \ \ \leq \ \ \ddot{\bld{x}}_{max}(\bld{x}_{k}) \label{const_d} \\[-2pt]
		\bld{0} & = & \bld{x}_{k+1} - \bld{x}_{k} - \Delta_t \dot{\bld{x}}_{k+1} \label{const_e} \\[-2pt]
		\bld{0} & = & \dot{\bld{x}}_{k+1} - \dot{\bld{x}}_{k} - \Delta_t \ddot{\bld{x}}_{k+1} \label{const_f}
	\end{eqnarray}
 \label{eq:costf2_}
 \end{subequations}
where $i$ and $j$ iterate over the four wheels' indexes; $\psi$ indicates the weighting factors; constraint (\ref{const_a}) ensures that the normal forces are positive in order to avoid loosing contact with the ground; $\bld{x}_{min}$, $\dot{\bld{x}}_{min}$, and $\ddot{\bld{x}}_{min}$ are the minimum boundaries for position, velocity, and acceleration of the linear actuators; and $\bld{x}_{max}$, $\dot{\bld{x}}_{max}$, and $\ddot{\bld{x}}_{max}$ are the maximum boundaries. Equations (\ref{const_e}) and (\ref{const_f}) ensure the kinematic consistency in velocity and acceleration, where $\Delta_t= t_{k+1}-t_{k}$. Please note that $\dot{\bld{x}}_{min}$, $\ddot{\bld{x}}_{min}$, $\dot{\bld{x}}_{max}$, and $\ddot{\bld{x}}_{max}$ are functions of $\bld{x}_{k}$, which are $\arctan$-based smooth functions to avoid the mechanical shock of the piston produced when achieving the position boundaries at high speed. This helps to reduce the induced mechanical disturbances.
\section{Numerical Simulations and Results}
\label{sec:results}
Let us simulate the inertial parameters and normal forces modeling and the optimal control described in previous sections on the heavy-duty mobile manipulator depicted in Fig. \ref{fig:platform} that contains four wheels, two linear actuators that power the articulated suspension and a mounted 7-DoF parallel-serial manipulator. This simulation on flat terrain was implemented on M{\scriptsize ATLAB} and using Simscape as physics engine and \textit{fmincon} as optimal nonlinear solver. A laptop endowed with standard capabilities, 2.40G [Hz], and 32 [GB] RAM was used.

First, we verify the accuracy and performance of the analytic wheels normal forces with the variable inertial parameters method described in Section \ref{sec:forces}. For this simulation we run the Simscape and analytic models for 100 [s] with fixed time step of $1e^{-3}$ [s] with the following commands
\begin{eqnarray}
    \hspace{-1.0cm}\mbox{Manipulator} & \bld{q}_{arm} & = \quad \bld{q}_{\alpha} + \bld{q}_{\beta}\sin(t) \label{eq:mani} \\
    \hspace{-1.0cm}\mbox{Wheels} & \dot{\bld{q}}_{w} & = \quad {\footnotesize \begin{bmatrix}
        \pi/5 & \pi/10 & \pi/5 & \pi/10
    \end{bmatrix}^{\top}} \label{eq:whee} \\
    \hspace{-1.0cm}\mbox{Suspension} & \bld{x} & = \quad \bld{x}_\alpha + \bld{x}_{\beta}\sin(f_s \ t)
\end{eqnarray}
where $\bld{q}_{\alpha} = {\footnotesize [ \pi/2 \ \, 0.5 \, -\!0.5 \ \, 0.1 \ \, 0 \ \, \pi/2 \ \, 0 ]^{\top}}$, $f_s \!\!=\!\! 0.3$, $\bld{x}_{\alpha} \!\!=\!\! {\footnotesize [ -\!0.1 \, -\!0.1 ]^{\top}}$, $\bld{q}_{\beta} = {\footnotesize [ 4\pi \ \, 0.7 \, -\!0.4 \ \, 0.9 \ \, 0 \ \, 0 \ \, 0 ]^{\top}}$ and $\bld{x}_{\beta} \!\!=\!\! {\footnotesize [ 7e^{-2} \ \, 7e^{-2} ]^{\top}}$. We show in Fig. \ref{fig:forces_k} the four wheels normal forces where the Simscape ground truth is retrieved by using the spatial contact force block of Simulink for the wheel ground interaction assuming a flat terrain. At the same time two analytic normal forces solutions are plotted, the fixed solution considers the mobile base a single rigid body with constant inertial parameters and the variable solution considers the mobile base as an articulated-actuated multibody mechanism where its inerial paratemers are variables. The accuracy of our approach can be verified in Fig. \ref{fig:forces_error} where the variable and fixed analytic solutions errors with respect to the Simscape solution are depicted. The errors in $f_{\!F\!L}^n$, $f_{\!F\!R}^n$, $f_{\!R\!L}^n$, $f_{\!R\!R}^n$ have RMS values of 171.4 [N], 228.2 [N], 376.6 [N] and 377.1 [N] for our variable solution and values of 2030.1 [N], 2036.8 [N], 2072.5 [N], 2082.6 [N] for the fixed solution. Therefore, the errors in our solution represent only 2.46 [\%], 3.7 [\%], 2.1 [\%] and 2.1 [\%] of inaccuracy for each wheel while the fixed solution reported in \cite{petrovic2022analytic} represents 20.2 [\%], 21.7 [\%], 12.2 [\%], 11.5 [\%] of inaccuracy.

\begin{figure}[h!]
	\centering
	\begin{minipage}[b]{0.45\textwidth}
		\includegraphics[trim={0.0cm 0.08cm 0.1cm 0.0cm},clip,width=\textwidth]{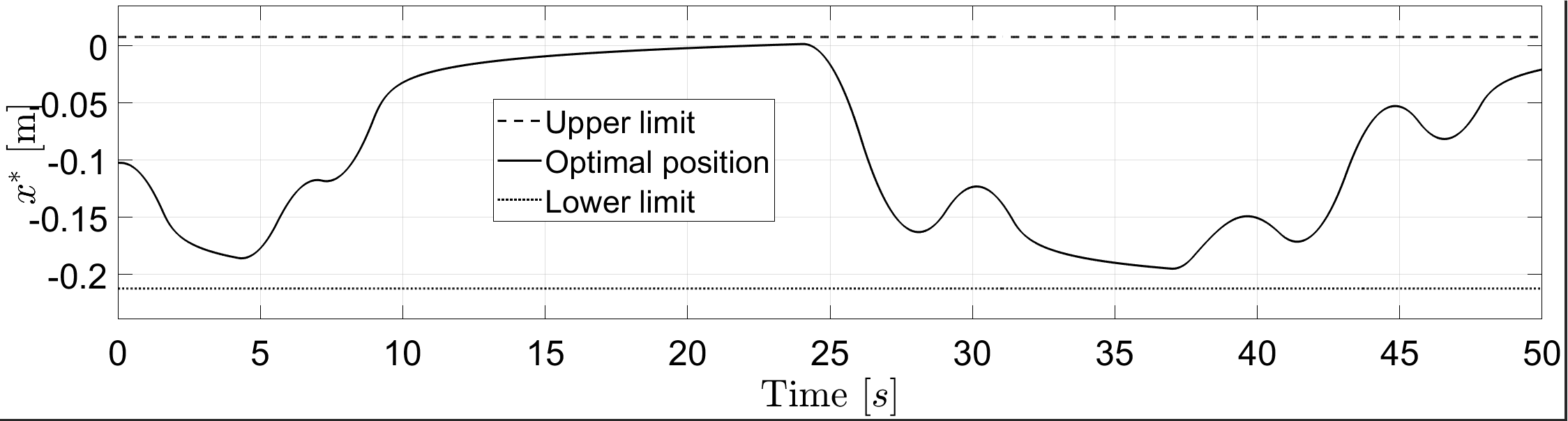}
	\end{minipage}
    
    \vspace{-0.0cm}
	\begin{minipage}[b]{0.45\textwidth}
		\includegraphics[trim={0.0cm 0.08cm 0.1cm 0.0cm},clip,width=\textwidth]{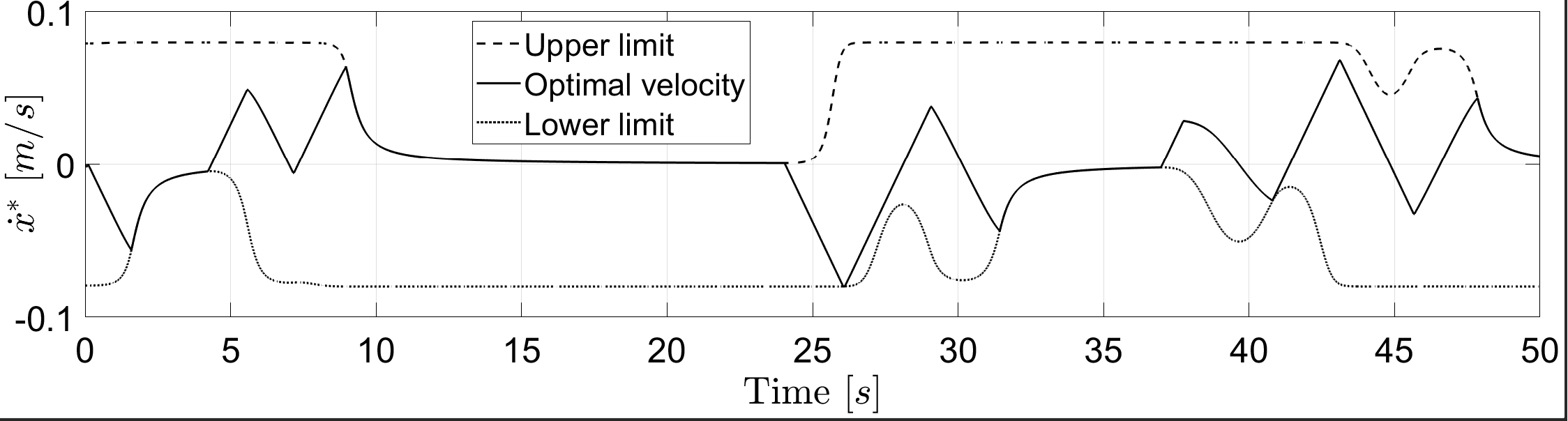}
	\end{minipage}

    \vspace{-0.0cm}
	\begin{minipage}[b]{0.45\textwidth}
		\includegraphics[trim={0.0cm 0.08cm 0.1cm 0.0cm},clip,width=\textwidth]{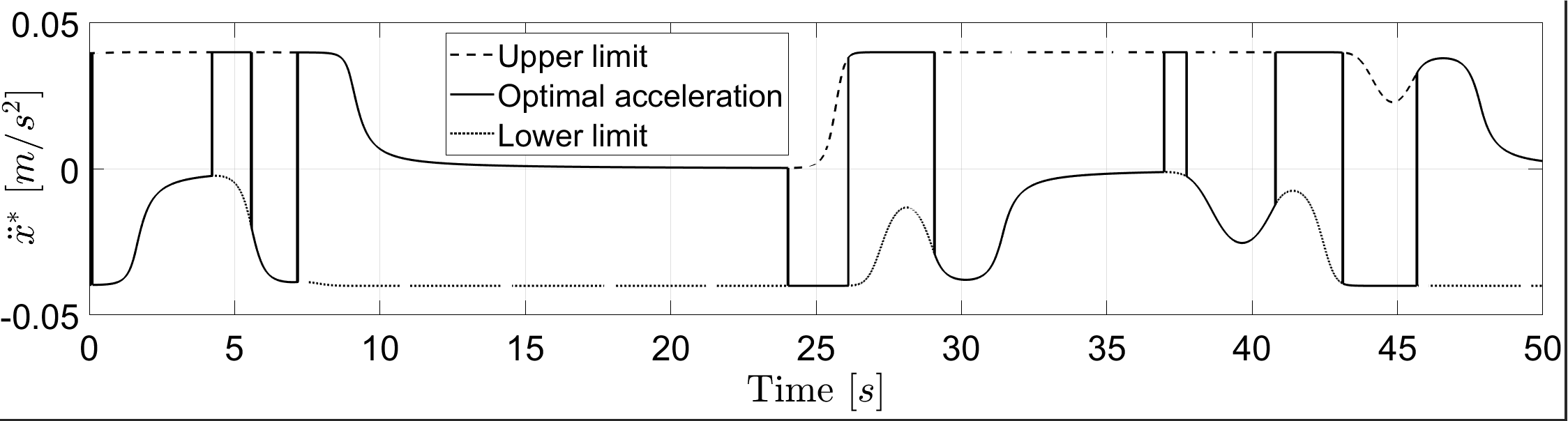}
	\end{minipage}

	\vspace*{-0.0cm}
	\caption{{\footnotesize {\bf Optimal position, velocity and acceleration of linear actuators through time.} Notice how the upper and lower boundaries of velocity and acceleration vary when the actuator position is approaching one of its limits.}}
	\label{fig:optimal}
\end{figure}

Additionally we test our numeric optimal control described by (\ref{eq:costf2}-\ref{eq:costf2_}), for this we set the same manipulator and wheels commands described by (\ref{eq:mani}) and (\ref{eq:whee}), and we assume that both actuators of the suspension have the same kinematics, it means $x \!=\! x_R \!=\! x_L$, $\dot{x} \!=\! \dot{x}_R \!=\! \dot{x}_L$ and $\ddot{x} \!=\! \ddot{x}_R \!=\! \ddot{x}_L$. The weighting factors $\psi_i$ are set equal to 1 for the difference among rear and front wheels' forces and 1$e^{-3}$ for the difference among lateral wheels' forces. The weights $\psi_\kappa$ are set equal to 100 for the rear and front stability angles and 10 for the lateral angles. Then, if the manipulator and wheels commands $\bld{q}_{arm}$ and $\bld{q}_{w}$ are known but independently controled, our optimal control finds the best next actuators state that better stabilizes the mobile manipulator in both dynamic and static manners. The wheel normal forces are depicted in Fig. \ref{fig:forces_control} where the fixed forces represent those normal forces when the mobile platform is not actuated and the controled forces refer to the ground reaction forces when the suspension is articulated, actuated and controled for stabilization. The optimal position, velocity and accelerations of the actuators in illustrated in Fig. \ref{fig:optimal} where the limit values are respected. As aforementioned the velocity and acceleration limits are variable to avoid the mechanical piston shock that induces considerable disturbances in the Simscape physics engine.

\begin{figure}[h!]
	\centering
	\begin{minipage}[b]{0.45\textwidth}
		\includegraphics[trim={0.0cm 0.08cm 0.1cm 0.0cm},clip,width=\textwidth]{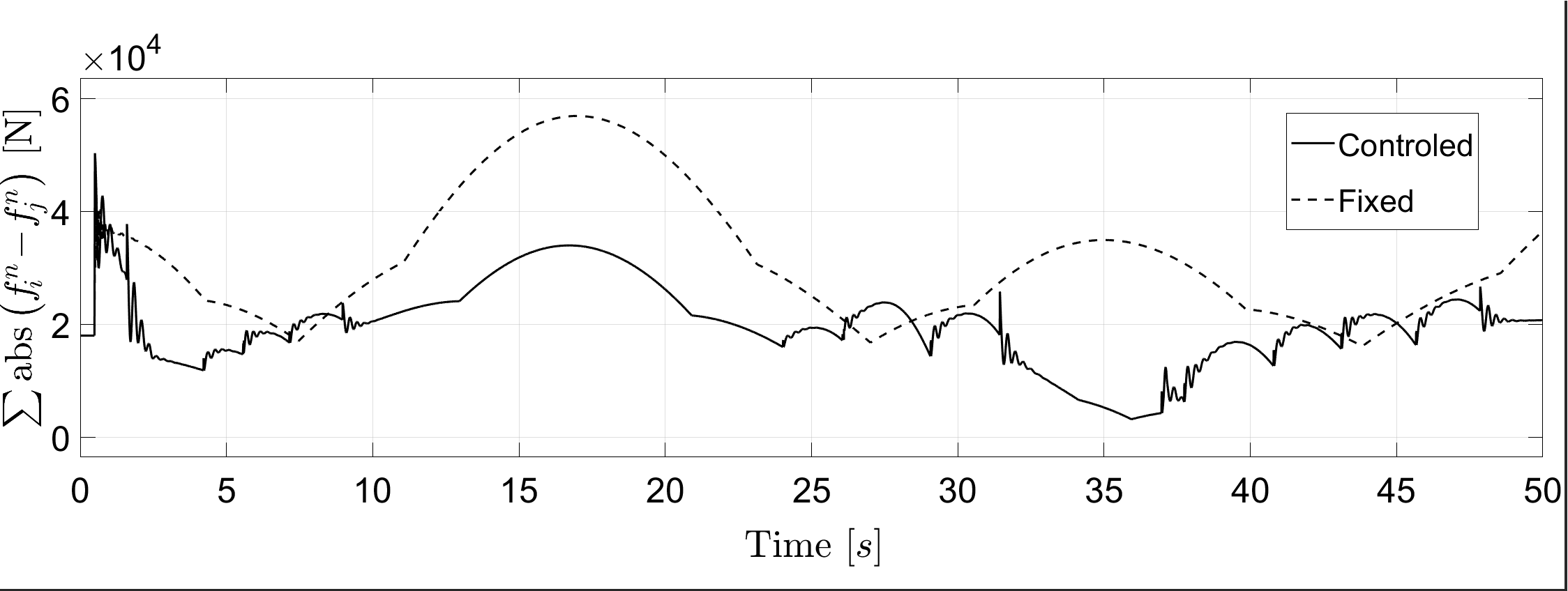}
	\end{minipage}
    \vspace*{-0.0cm}
	\caption{{\footnotesize {\bf Metric for normal forces distribution.} According to the first criteria of the cost function (\ref{eq:costf2}), the normal forces distribution is optimized by penalizing the difference among them. The lower the absolute sum the more uniformly the normal forces are distributed on the wheels.}}
	\label{fig:diff_forces}
\end{figure}
\begin{figure}[h!]
	\centering
	\begin{minipage}[b]{0.45\textwidth}
		\includegraphics[trim={0.0cm 0.08cm 0.1cm 0.0cm},clip,width=\textwidth]{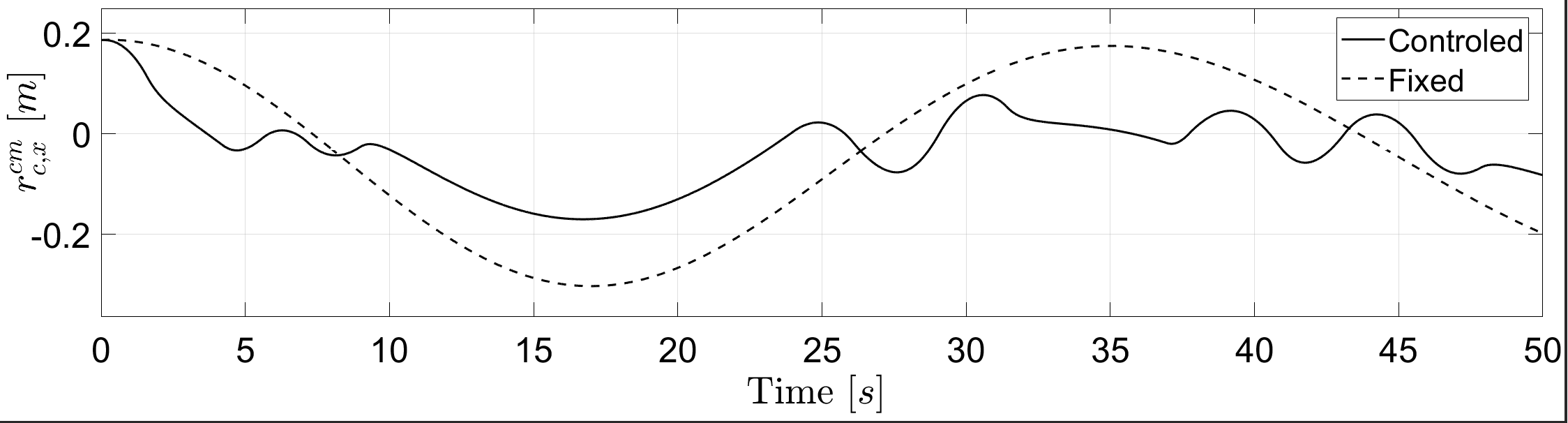}
	\end{minipage}
    \vspace*{-0.0cm}
	\caption{{\footnotesize {\bf Mobile manipulator's center of mass.} We depict its position in $x$-axis expressed at chassis frame $\Sigma_c$. According to the second criteria of the cost (\ref{eq:costf2}), the center of mass is stabilized by minimizing its distance from the chassis frame $\Sigma_c$. The closer to zero the more statically stable.}}
	\label{fig:com}
\end{figure}

We can now set two metrics to measure the dynamic and static stabilization performance of our optimal control. For the dynamic stabilization we define the following metric as the difference among all wheels normal forces to measure the normal forces distribution
\begin{equation}
    \sum_{i,j} \mbox{abs} \left( f_{\!i}^n \!-\! f_{\!j}^n \right) \quad \forall \ \{ _{F\!R}, _{F\!L}, _{R\!R}, _{R\!L} \}, \quad i\neq j
\end{equation}
The results of this metric is shown in Fig. \ref{fig:diff_forces} where the normal forces distribution is more uniform in our optimal control than the distribution when the suspension is fixed. For this simulation our controled solution has a RMS value of 2.1$e^{4}$ [$N$] in comparison with 3.3$e^{4}$ [$N$] of the fixed solution. It means the lower the metric the better stabilization due to a better normal forces distribution.

On the other hand, we consider a metric for the static stability based on the mobile manipulator center of mass position, with respect to the chassis frame, which is directly related to the stability margination metric reported in \cite{iagnemma2003control}. We show in Fig. \ref{fig:com} the center of mass position in its $x$ axis when the suspension is controled and fixed. In this case the metric is better when it is closer to zero, it means the center of mass projection onto the terrain is closer to the origin of the chassis frame, which from Fig. \ref{fig:snaps} we realize it is the center of the polygon of support generated by the four-wheels contact points. For this metric through time our controled solution has a RMS value of 8$e^{-2}$ [$m$] while the fixed solution has a value of 16.5$e^{-2}$ [$m$].

\begin{figure}[h!]
	\centering
	\begin{minipage}[b]{0.45\textwidth}
		\includegraphics[trim={0.0cm 0.08cm 0.1cm 0.0cm},clip,width=\textwidth]{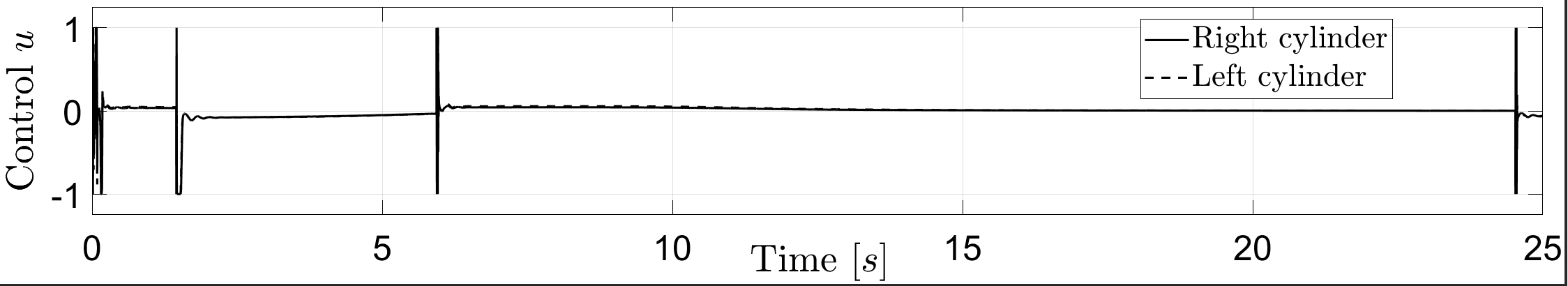}
	\end{minipage}
    
    \vspace{-0.0cm}
	\begin{minipage}[b]{0.45\textwidth}
		\includegraphics[trim={0.0cm 0.08cm 0.1cm 0.0cm},clip,width=\textwidth]{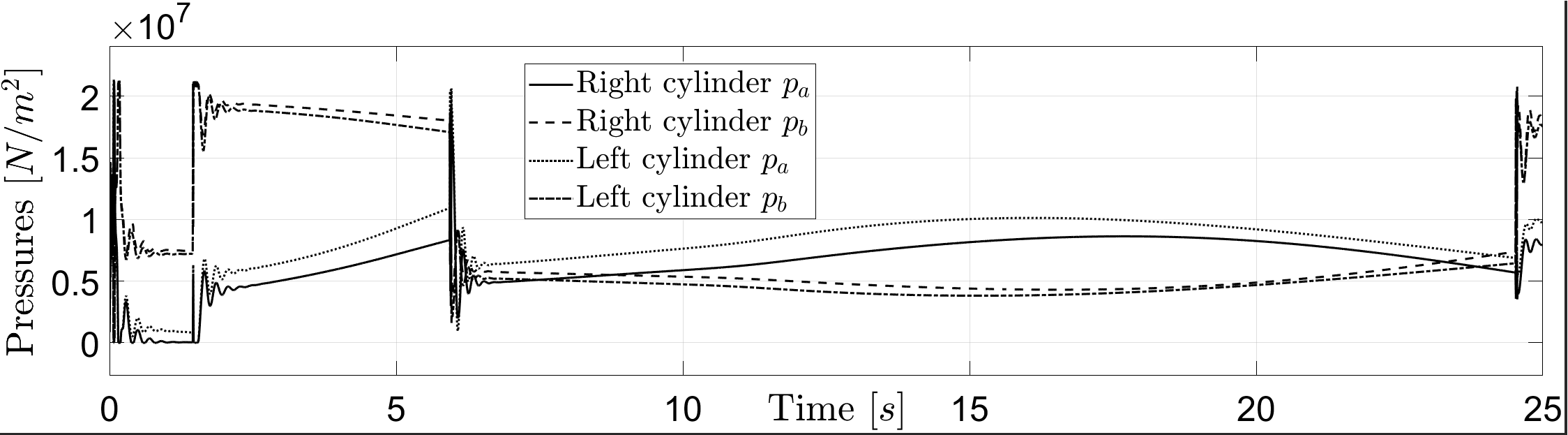}
	\end{minipage}

    \vspace{-0.0cm}
	\begin{minipage}[b]{0.45\textwidth}
		\includegraphics[trim={0.0cm 0.08cm 0.1cm 0.0cm},clip,width=\textwidth]{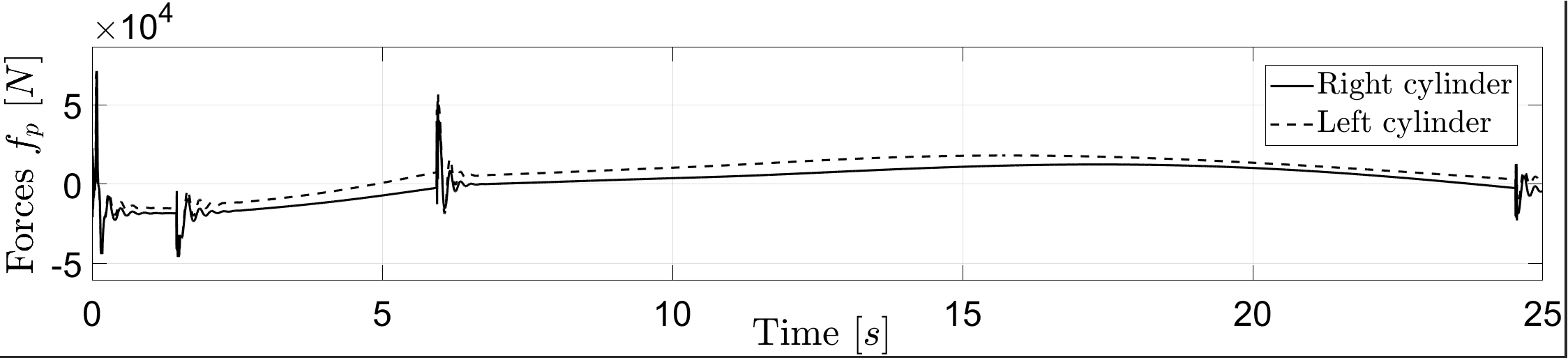}
	\end{minipage}

    \vspace{-0.0cm}
	\begin{minipage}[b]{0.45\textwidth}
		\includegraphics[trim={0.0cm 0.08cm 0.1cm 0.0cm},clip,width=\textwidth]{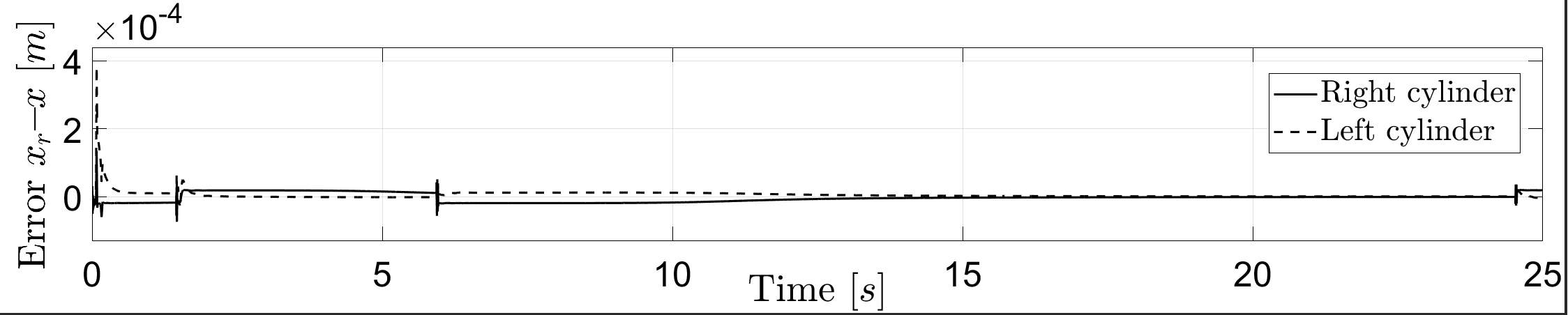}
	\end{minipage}

	\vspace*{-0.0cm}
	\caption{{\footnotesize {\bf Hydraulic valve control for suspension actuators.} For right and left side actuators in the suspension are shown, from top to bottom: the valve control signal $u$, pressures in the chambers, delivered force by the piston and tracking error in position through time.}}
	\label{fig:hydra}
\end{figure}

If we consider the optimal actuators' position as a reference trajectory $x_r(t)$ for the hydraulic system presented in Section \ref{sec:hydra}, then the valve control $u$ can be written as a proportional control law
\begin{equation}
    u \ = \ k_p (x_r(t) - x)
    \label{sss}
\end{equation}
which can be used as a low-level controller for the suspension actuators. Let us consider $x$ in (\ref{sss}) as the measured position from sensors. Thus, regarding a Bosch Rexroth valve parameters 40 l/min\@$\Delta p$=3.5 [MPa], a 25 [s] valve control simulation, with $k_p = 10$ for both actuators, is performed and the results are depicted in Fig. \ref{fig:hydra}, where the control efficiency can be noted in the position error.
\begin{table}[!t]
\centering
\begin{tabular}{|c|c|}
\hline
{\bf Mechanical expression} & {\bf Time} \\
\hline
Forward kinematics from $\Sigma_w$ to wheels' frames & 0.47 [$\mu s$] \\
Manipulator's total exerted wrench $\F_{\!m}$ & 2.7 [$\mu s$] \\
Manipulator's inverse dynamics & 2.8 [$\mu s$] \\
Inertial parameters $m_w$, $\bld{r}^{\tiny\mbox{cm}}_w$, $\bld{\mathcal{I}}_w$ & 5.2 [$\mu s$] \\
Wheels' normal forces $f_{\!F\!R}^n$, $f_{\!F\!L}^n$, $f_{\!R\!R}^n$, and $f_{\!R\!L}^n$ & 10.0 [$\mu s$] \\
\hline
\end{tabular}
\vspace{0.2cm}
\caption{{\footnotesize {\bf C++ computational times in microseconds.} \textnormal{These times correspond to the average for a single execution of the corresponding expressions.}}\label{tab:table1}}
\end{table}

In addition, we implemented C++ routines for testing the computation time for our analytic solutions for several kinematic and dynamic expressions. The time is measured in microseconds and the values reported in TABLE \ref{tab:table1} are the average time when executing the corresponding algorithms 1e6 times when input commands for wheels, manipulator and suspension are provided.
\section{Conclusions}
In this work we have provided a complete formulation for modeling and control of non redundant active articulated suspension of heavy-duty mobile manipulators. We successfully combine in a single procedure explicit analytical solutions for the closed kinematic chains mechanisms in the suspension, the manipulator's dynamics and the assembly inertias method for enhancing the wheels normal forces accuracy computation. As a result we obtain an average inaccuracy error of 2.6 [$\%$] in comparison with the 16.4 [$\%$] error retrieved with the constant inertial-parameters solution reported in \cite{petrovic2022analytic}. Moreover, we tested our inertial parameters and normal forces solutions into a multiobjective optimal control for static and dynamic stabilization of the mobile manipulator in flat terrain. By implementing our control, the active suspension was capable to reduce from 3.3$e^{4}$ [$N$] to 2.1$e^{4}$ the difference among normal forces, which provides a better normal forces distribution, and from 16.5$e^{-2}$ [$m$] to 8$e^{-2}$ [$m$] the center of mass position from the center of the polygon of support, which provides better static stabilization. Our approach also mitigates the piston mechanical shock by including smooth joint limits function. As we have shown the C++ fast computational times through TABLE \ref{tab:table1} of our algorithms, we consider, as future work, that this active suspension model and control can be suitable for fast and accurate computation for real-time online overturning-free optimal motion generation for rough-terrain traversing since the well-known nonlinear model predictive control approaches \cite{diehl2009efficient,diehl2006fast} require to evaluate the robot's dynamics, in an accurate and fast manner, several times per iteration of the state feedback control.

\bibliographystyle{IEEEtran}

	\bibliography{jrnl_root}

\begin{IEEEbiography}[{\includegraphics[width=1in,height=1.25in,clip,keepaspectratio]{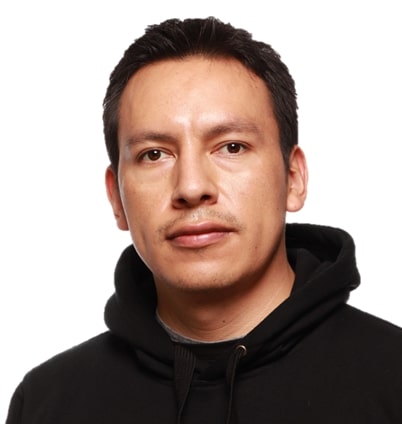}}]{Álvaro Paz Anaya} obtained his bachelor's degree in Electronics Engineering in 2013 from Instituto Tecnológico de San Juan del Río, Querétaro, Mexico. He received the M.Sc. and PhD degrees in Robotics and Advanced Manufacturing from Centro de Investigación y de Estudios Avanzados del IPN (CINVESTAV), Saltillo, Mexico in 2017 and 2022. He is currently a post doctoral researcher at Tampere University, Tampere, Finland. His research interests include robot trajectory optimization, humanoid whole body motion generation, multibody dynamic algorithms and mobile robotics.
\end{IEEEbiography}

\begin{IEEEbiography}[{\includegraphics[width=1in,height=1.25in,clip,keepaspectratio]{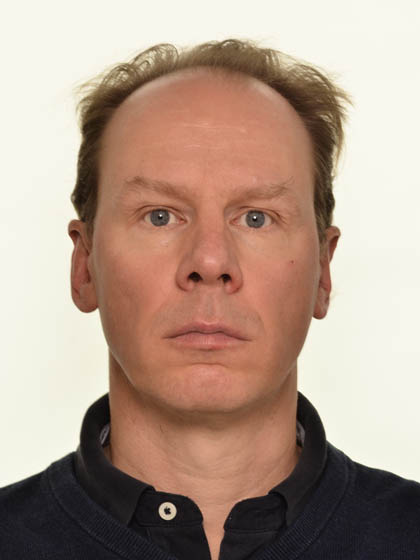}}]{Jouni Mattila}
received an M.Sc. and Ph.D. in automation engineering from Tampere University of Technology, Tampere, Finland, in 1995 and 2000, respectively. He is currently a professor of machine automation with the Unit of Automation Technology and Mechanical Engineering at Tampere University. His research interests include machine automation, nonlinear-model-based control of robotic manipulators, and energy-efficient control of heavy-duty mobile manipulators.
\end{IEEEbiography}

 





\end{document}